\documentclass[lettersize,journal]{IEEEtran}
\usepackage{amsmath,amsfonts}
\usepackage{algorithmic}
\usepackage[ruled,linesnumbered]{algorithm2e}
\usepackage{array}
\usepackage[caption=false,font=normalsize,labelfont=sf,textfont=sf]{subfig}
\usepackage{picinpar}
\usepackage{textcomp}
\usepackage{stfloats}
\usepackage{caption}
\usepackage{url}
\usepackage{float}
\usepackage{hyperref}
\usepackage{graphicx}
\usepackage{flushend}
\usepackage{colortbl}
\usepackage{soul}
\usepackage{multirow}
\usepackage{pifont}
\usepackage{alltt}
\usepackage{epstopdf}
\usepackage{dsfont}
\usepackage{cite}
\usepackage{xcolor}
\usepackage{setspace} 
\newtheorem{definition}{\rm{\textbf{Definition}}} 
\newtheorem{theorem}{\rm {\textbf{Theorem}}}
\newcommand{\white}{\textcolor{white}}

\usepackage{array,tabularx,booktabs}

\begin{document}
\begin{spacing}{0.98} 
\title{\huge Semi-Supervised Federated Learning via Dual Contrastive Learning and Soft Labeling for Intelligent Fault Diagnosis}

\author{
	\vskip 1em
	
	Yajiao Dai, Jun Li,~\IEEEmembership{Fellow,~IEEE,} Zhen Mei,~\IEEEmembership{Member,~IEEE,} Yiyang Ni,~\IEEEmembership{Senior Member,~IEEE,} 
    
    Shi Jin,~\IEEEmembership{Fellow,~IEEE,}  Zengxiang Li, Sheng Guo,~\IEEEmembership{Member,~IEEE,} Wei Xiang,~\IEEEmembership{Senior Member,~IEEE.}

    \thanks{This work was supported (except for Wei Xiang) in part by the Key Technologies R\&D Program of Jiangsu (Prospective and Key Technologies for Industry) under Grants BE2023022 and BE2023022-2; in part by National Natural Science Foundation of China (NSFC) under Grant 62471204; in part by Major Natural Science Foundation of the Higher Education Institutions of Jiangsu Province under Grant 24KJA510003; in part by the National Natural Science Foundation of China under Grants 62201258; and in part by “the Fundamental Research Funds for the Central Universities", No.30923011035. (Corresponding authors: Jun Li; Zhen Mei.)}
	\thanks{
	
		Y. Dai and Z. Mei are with the School of Electronic and Optical Engineering, Nanjing University of Science and Technology, Nanjing 210094, China (e-mail: \{yajiao.dai, meizhen\}@njust.edu.cn).
		
		J. Li and S. Jin are with the School of Information Science and Engineering, Southeast University, Nanjing, 210096, China (e-mail: \ jun.li, jinshi@seu.edu.cn).
		
		Y. Ni is with the  Institute of Artificial Intelligence Research, Jiangsu Second Normal University, Nanjing 210013, China, and also with Jiangsu Key Laboratory of Wireless Communications, Nanjing University of Posts and Telecommunications, Nanjing 210003, China (e-mail: niyy@jssnu.edu.cn).

        Z. Li and S. Guo are with the Digital Research Institute, ENN Group, Langfang, 065001, China. (e-mail: \{lizengxiang,guoshengd\}@enn.cn).

        W. Xiang is with the School of Computing, Engineering and Mathematical Sciences, La Trobe University, Melbourne, VIC 3086, Australia, and also with the College of Science and Engineering, James Cook University, Cairns, QLD 4878, Australia (e-mail: w.xiang@latrobe.edu.au).
	}
}
\maketitle

\begin{abstract}
Intelligent fault diagnosis (IFD) plays a crucial role in ensuring the safe operation of industrial machinery and improving production efficiency. However, traditional supervised deep learning methods require a large amount of training data and labels, which are often located in different clients. Additionally, the cost of data labeling is high, making labels difficult to acquire. Meanwhile, differences in data distribution among clients may also hinder the model's performance. To tackle these challenges, this paper proposes a semi-supervised federated learning framework, SSFL-DCSL, which integrates dual contrastive loss and soft labeling to address data and label scarcity for distributed clients with few labeled samples while safeguarding user privacy. It enables representation learning using unlabeled data on the client side and facilitates joint learning among clients through prototypes, thereby achieving mutual knowledge sharing and preventing local model divergence. Specifically, first, a sample weighting function based on the Laplace distribution is designed to alleviate bias caused by low confidence in pseudo labels during the semi-supervised training process. Second, a dual contrastive loss is introduced to mitigate model divergence caused by different data distributions, comprising local contrastive loss and global contrastive loss. Third, local prototypes are aggregated on the server with weighted averaging and updated with momentum to share knowledge among clients. To evaluate the proposed SSFL-DCSL framework, experiments are conducted on two publicly available datasets and a dataset collected on motors from the factory. In the most challenging task, where only 10\% of the data are labeled, the proposed SSFL-DCSL can improve accuracy by 1.15\% to 7.85\% over state-of-the-art methods.  
\end{abstract}


{}


\section{Introduction}

    \IEEEPARstart{W}{ith} the rapid advancement of modern industry, the complexity and intelligence of industrial equipment have increased dramatically, making intelligent fault diagnosis (IFD) crucial. IFD is vital to ensure equipment operation, improve production efficiency, and support technological and socio-economic advancement in industry \cite{iot9745085}.

    In recent years, numerous deep learning (DL)-based IFD methods have emerged using extensive data from factories \cite{9796011,9994749,10043805,10287861,DBLP:journals/ieeejas/YangLLLN24,DBLP:journals/ieeejas/RenWZYCN24,fedavg,ChenLYYG23,DBLP:journals/tits/ZhangLSDNTWH24}. These methods mainly rely on supervised learning, which requires many labeled samples to achieve stable convergence. However, acquiring high-quality labeled data is challenging, often necessitating significant resources and risking privacy breaches, especially as industrial equipment is distributed across various locations, creating data islands. Furthermore, due to differences in equipment, operating environments, and data collection methods, as well as variations in the likelihood of failures and differences in how data are collected, the data distribution becomes diversified. These will degrade the generalizability of the models.

    To address the data islands problem in IFD, federated learning (FL) \cite{iot9718548,GengHLL22,LiSWDMSHP22,10309847,DBLP:journals/corr/abs-2202-09027,10371403,iot9548946,10122855}  has been adopted to enable distributed collaborative training while preserving client data privacy. In \cite{GengHLL22}, aggregation was improved by incorporating the F1 score of each client's classification task. A lightweight FL framework was proposed in \cite{10309847}, in which computation and communication costs are reduced, making it suitable for edge devices. In \cite{iot9548946}, a stacking model was integrated with FL and enhanced particle swarm optimization was designed to address data imbalance and contamination, thereby improving the robustness of IFD. Additionally, in \cite{10122855}, the data islands issue was mitigated by locally training convolutional autoencoders and aggregating them into a global model, enhancing IFD performance. However, these approaches still struggle with insufficient labeling and data distribution imbalances in practical settings.

    Recent FL research has addressed data heterogeneity through federated transfer learning (FTL) and federated meta-learning (FML). For instance, a domain-adversarial network was employed in \cite{ChenHLLGL20} to align data distributions for cross-domain IFD. At the same time, a weighted FTL framework was developed in \cite{ChenLHYCL22} to construct a robust, privacy-preserving global diagnostic model. In \cite{iot10669849}, a federated generalized zero-sample IFD paradigm was proposed by combining FL and zero-shot learning, enabling both seen and unseen faults to be diagnosed without data sharing. An asynchronous domain adaptation FL architecture was designed in \cite{DBLP:journals/kbs/WanNLLL24}, where a multi-perspective distribution discrepancy aggregation strategy was used to mitigate negative transfer. Additionally, in \cite{ChenTL23}, an FML approach was introduced to facilitate rapid adaptation in few-shot IFD scenarios using indirectly owned datasets.  Despite these advances, these methods still fail to exploit the abundant unlabeled industrial data.

\begin{table*}[ht]
  \centering
  \renewcommand{\arraystretch}{1.5}   
  \caption{COMPARISION OF FEDERATED LEARNING METHODS}
  \label{tab:scenario_fl_comparison}
  {\fontsize{8.5}{8}\selectfont
  \resizebox{\textwidth}{!}{
  \begin{tabular}{
    >{\raggedright\arraybackslash}p{2.4cm}|
    >{\raggedright\arraybackslash}p{1.9cm}|
    >{\raggedright\arraybackslash}p{1.7cm}|
    >{\raggedright\arraybackslash}p{2.1cm}|
    >{\raggedright\arraybackslash}p{6.2cm}}
    \hline\hline
    \textbf{Scenario} & \textbf{Label Setting} & \textbf{Heterogeneity} & \textbf{Communications} & \textbf{SSFL-DCSL Advantages} \\
    \hline
    Privacy \cite{iot9718548,GengHLL22,LiSWDMSHP22,10309847,DBLP:journals/corr/abs-2202-09027,10371403,iot9548946,10122855}  & All labeled & Partial & Moderate & 
    Enables joint learning with labeled and unlabeled data via prototype-only aggregation for enhanced privacy. \\
    \hline
    Data heterogeneity \cite{ChenHLLGL20,ChenLHYCL22,iot10669849,DBLP:journals/kbs/WanNLLL24,ChenTL23}  & All labeled & Local/meta & Moderate & 
    Achieves robust local-global alignment through dual-level prototype contrast and momentum stabilization. \\
    \hline
    Label scarcity \cite{RizveDRS21,SajjadiJT16,ZhangSHHW22,YangTYDZ24,9964271,9999706,10100629}  & Few labeled + pseudo-labels & Consistency & High & 
    Mitigates pseudo-label noise and improves data efficiency using Laplace-weighted pseudo-labeling and contrastive prototype learning. \\
    \hline \hline
  \end{tabular}
  }
  }
  \label{diff-fl}
\end{table*}

    Semi-supervised learning (SSL) effectively leverages both labeled and unlabeled data to improve feature extraction and generalization \cite{RizveDRS21,SajjadiJT16,ZhangSHHW22,YangTYDZ24,9964271,9999706,10100629}. Techniques like pseudo-labeling \cite{RizveDRS21} and consistency regularization \cite{SajjadiJT16} enhance neural network generalization. In \cite{ZhangSHHW22}, Monte Carlo-based adaptive threshold selection was utilized to enable semi-supervised IFD with limited labels, while a contrastive image generation network with multimodal data was developed in \cite{YangTYDZ24} for real-time IFD. A method combining weighted label propagation and virtual adversarial training was introduced in \cite{9964271} for gearbox IFD. In \cite{9999706}, a meta self-training framework was proposed, where pseudo-labels generated by a teacher network were used to guide the training of a student network. Additionally, an online SSL contrastive graph generative network was presented in \cite{10100629} for harmonic drives, leveraging multimodal data. Although these methods have shown considerable success, they are all based on centralized training, and knowledge sharing of unlabeled data among users is not sufficiently mined.

    \subsection{Motivation and Contribution}
    
    Despite recent advancements, three key challenges limit the deployment of data-driven fault diagnosis models in real-world factories:
    
    1) Data islands and privacy: Sensor data are naturally distributed across geographically and administratively separate clients, making centralized training infeasible without risking proprietary information leakage.
    
    2) Label scarcity: Accurate labeling of fault data requires expert knowledge and expensive downtime, so most industrial datasets contain very few annotations.
    
    3) Data heterogeneity: Variations in equipment types, operating conditions, and sampling setups lead to highly non-IID data distributions that degrade model generalization.

    Modern factories, however, naturally produce vast streams of unlabeled sensor data, an untapped asset if we can overcome privacy and heterogeneity barriers. Motivated by the vision of turning this “unlabeled abundance” into stronger, more robust diagnostic models, we propose to fuse client-side self-supervised representation learning with lightweight, prototype-level knowledge exchange.

    Accordingly, we introduce SSFL-DCSL, a semi-supervised federated-learning framework that integrates soft-label weighting and a dual contrastive loss to empower edge clients with only a handful of labels. Our method enables each client to refine its representations via self-supervision on unlabeled data while sharing compact prototypes globally to align disparate feature spaces.

    Table \ref{diff-fl} summarizes the differences between the existing federated learning methods and the proposed method. The main contributions are summarized as follows.

    1) We introduce SSFL-DCSL, an approach that integrates SSL based on pseudo labels, prototypes-based knowledge sharing, consistency regularization, and FL to facilitate accurate and efficient distributed model training for IFD to address issues including limited labels, inconsistent data distribution, and privacy leakage.

    2) To counteract the adverse effects of unreliable pseudo-labels on model performance, we develop a truncated Laplace-based adaptive sample weighting (TLAW) function. This function assigns weights to samples with low-quality pseudo labels according to confidence deviations, safeguarding the training process from pseudo-label inaccuracies, and enhancing overall model performance.

    3)  To fully utilize abundant unlabeled data and facilitate global knowledge sharing, a dual contrastive loss is adopted. The local contrastive loss (LCL) drives the model to learn robust representations via self-supervised consistency between augmented views, while the global contrastive loss (GCL) aligns local features with global prototypes, addressing data heterogeneity and enhancing inter-client knowledge transfer.

    4) To mitigate the impact of sample heterogeneity and balance the model's performance across different categories, we employ a prototype aggregation method (PTA). Local prototypes are weighted and aggregated on the server and updated with momentum to facilitate knowledge sharing among clients.

    Comprehensive experiments are conducted using our SSFL-DCSL framework on three public datasets and a dataset collected from a partner factory. The results show that our method surpasses other baseline methods in IFD accuracy, with the SSFL-DCSL method increasing the accuracy from 1.15\% to 7.85\% in scenarios where only 10\% of the data are labeled.

    \begin{table*}[ht]
    \centering
    \small
    \setlength{\tabcolsep}{6pt}         
    \renewcommand{\arraystretch}{1}   
    \caption{KEY DIFFERENCES vs.\ PRIOR SSFL METHODS}
    \label{tab:ssfl-comparison}
    \begin{tabularx}{\textwidth}{
        >{\centering\arraybackslash}m{2.5cm}
        |>{\centering\arraybackslash}m{6.5cm}
        |>{\centering\arraybackslash}m{5.5cm}
        |>{\centering\arraybackslash}X
      }
    \hline \hline
    \textbf{Method} 
      & \textbf{Pseudo-label Filtering} 
      & \textbf{Heterogeneity Mitigation} 
      & \textbf{Aggregation} \\ 
    \hline
    
    FedMatch/FedProx \cite{FixMatch,fedprox}
      & Hard threshold $\rightarrow$ discard low-confidence samples
      & None 
      & Full model \\ 
    \hline
    
    FedCon\cite{fedcon}
      & No weighting (all pseudo-labels equally)
      & Contrastive on instances
      & Full model \\ 
    \hline
    
    FedCD\cite{liu2024fedcd}
      & Teacher-student consistency
      & Class-aware distillation
      & Full model \\ 
    \hline
    
    RSCFed\cite{RSCFed}
      & No explicit pseudo-label control
      & Random sampling consensus
      & Full model \\ 
    \hline
    
     \white{h}FedDB/(FL)\(^2\)  
    \cite{DBLP:conf/ijcai/Zhu0W0T0S24,DBLP:conf/nips/LeeL0L24}
      & Bias correction via Bayesian estimation
      & Consistency regularization
      & Full model \\ 
    \hline
    
    \textbf{Proposed SSFL-DCSL}
      & Weighting from Laplace distribution 
      & Local sample-level and global prototype-level alignment
      & Momentum Prototypes \\
    \hline \hline
    \end{tabularx}
    \label{diff-ssfl}
    \end{table*}

\subsection{Related Works}
    \subsubsection{Federated Learning for Industrial Fault Diagnosis} FL has emerged as a promising approach to collaboratively train models across distributed industrial systems without compromising data privacy \cite{iot9718548,GengHLL22}. For instance, efforts have focused on robust aggregation strategies tailored to industrial IoT \cite{iot9718548}, adaptive privacy preservation for maritime applications using homomorphic encryption \cite{iot9548946}, and discrepancy-based averaging to address domain shifts in rotating machinery diagnosis \cite{tim9789131}. Decentralization and trust enhancement were advanced through blockchain-assisted consensus frameworks \cite{LiSWDMSHP22}, while edge–cloud integration further improved training efficiency in motor fault diagnosis \cite{10122855}. To generalize across unseen classes, zero-sample federated frameworks synthesized fault types without data sharing \cite{iot10669849}, and federated distillation with GANs facilitated robust knowledge transfer under heterogeneous data distributions \cite{iot10891176}. Gradient alignment was explored to reduce negative transfer in rotating machinery diagnostics \cite{iot10949604}, and recent work introduced customized client-side aggregation for heterogeneous federated settings \cite{iot10742072}. 
    However, unlike our proposed SSFL-DCSL, existing FL-based industrial fault diagnosis methods do not explicitly address insufficient labeled data at local users, potentially limiting their effectiveness in practical scenarios.

    \subsubsection{Semi-Supervised Federated Learning} To address the scarcity of labeled data in federated settings, several SSFL frameworks have emerged. FedCon integrated instance-level contrastive learning to enrich feature representations under limited labels \cite{fedcon}. FedMatch and FedProx leveraged consistency regularization \cite{FixMatch} and unsupervised data augmentation (UDA) \cite{UDA}, respectively, enhancing robustness in non-IID environments \cite{fedprox}. FedCD employed dual-teacher architectures with class-aware distillation to mitigate class imbalance \cite{liu2024fedcd}, while RSCFed utilized random sampling consensus to ensure consistency across heterogeneous clients \cite{RSCFed}. Personalized federated learning with differential privacy tailored models to individual client distributions \cite{fedprox}. Recent methods directly tackled class prior bias through Bayesian debiasing (FedDB) \cite{DBLP:conf/ijcai/Zhu0W0T0S24}, and addressed confirmation bias with sharpness-aware consistency regularization ($\mathrm{FL})^2$ \cite{DBLP:conf/nips/LeeL0L24}. A systematic survey further categorized SSFL methodologies, summarizing their strengths and challenges \cite{DBLP:conf/ijcai/SongYZ0XK24}.
    Despite these advancements, effectively utilizing unlabeled data in highly heterogeneous federated environments without bias or instability remains challenging. The differences between the proposed SSFL-DCSL method and previous methods are detailed in Table \ref{diff-ssfl}.

     \subsubsection{Contrastive Learning for Fault Diagnosis} CL has been widely used to obtain discriminative features for IFD. Initially, semi-supervised multiscale permutation entropy-enhanced CL leveraged entropy-based positive pairs to detect subtle faults in rotating machinery with limited labeled data \cite{tim10203049}. Extending this idea, semi-supervised contrastive domain adaptation aligned feature distributions across conditions via adversarial and supervised contrastive objectives \cite{iot10909075}. For cross-domain tasks, dual-reweighted Siamese frameworks like DRSC mitigated multi-source domain imbalance by learning domain-invariant embeddings \cite{iot10944708}, while open-set methods employed synthesized negative samples to robustly detect novel anomalies \cite{tiiPeng2023OpenSetFD}. Supervised contrastive knowledge distillation was proposed for few-shot and incremental scenarios, preserving past representations and integrating new faults effectively \cite{tii10922759}. Domain discrepancy-guided contrastive learning selected hard positives and negatives to boost few-shot adaptation across varying conditions \cite{tii10032199}. Transformer-based methods further explored multimodal cross-sensor CL \cite{tim10989640}, and imbalance-aware CL adapted sampling strategies to address unevenly distributed fault data \cite{10151782}.
    However, despite promising results in centralized contexts, their potential in distributed and federated settings, especially concerning global-local knowledge sharing, remains largely unexplored.
    
    \white{\subsubsection{.}} The remainder of this article is organized as follows. Section \ref{2} outlines the system model. Section \ref{3} details the framework and key parts of the proposed SSFL-DCSL method. The experimental results and discussions are presented in Section \ref{4}, followed by the conclusion in Section \ref{5}.

 \begin{figure*}
 	\centering
 	\includegraphics[width=0.9\linewidth]{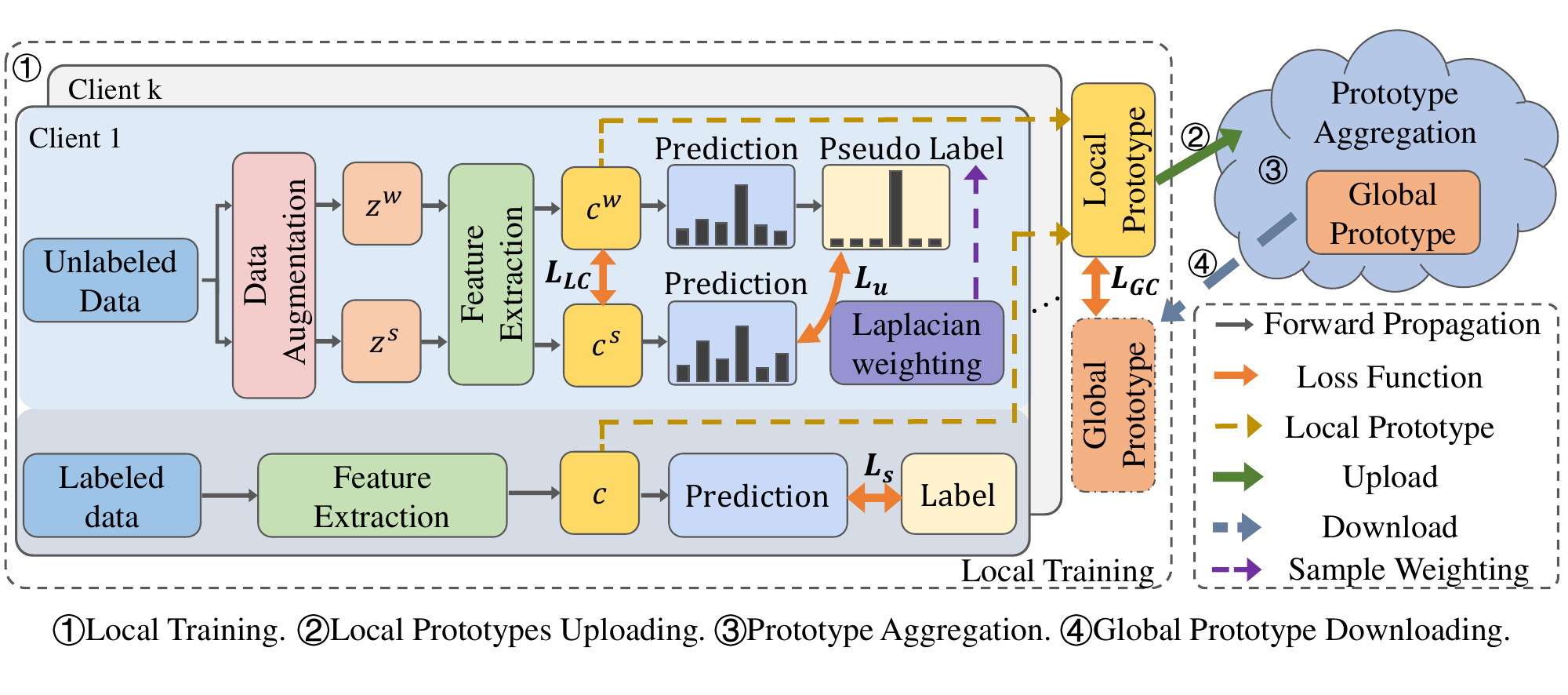}
    \vspace{-13pt}
 	\caption{\textrm{The overall framework of the local training stage of SSFL-DCSL.}}
 	\label{fig1}
 \end{figure*}

\section{System Model}\label{2}

    FL is designed to collaboratively develop a global model $G$, via coordinated communication across multiple clients. Each client maintains a local model denoted by $\mathcal{L} = \{l_{{\boldsymbol{\theta}}_k}\}_{k=1}^K$, where $K$ is the total number of users. Meanwhile, each client manages a part of the dataset $\mathcal{D} = \{\boldsymbol{x}_i, {\boldsymbol y}_i\}_{i=1}^N$, where $N$ is the number of samples, $\boldsymbol{x}_i$ represents the $i$-th training instance, and ${\boldsymbol y}_i$ is the corresponding one-hot label for the $C$ class classification problem, where $C$ is the number of categories. This dataset is divided into $K$ private sub-datasets $\mathcal{D}^k, k=1,2,...,K$, each owned by a different client, and the number of samples in $\mathcal{D}^k$ is ${N^k}$.

    In each round of communication, the global model $G$ initializes the local models ${{{\boldsymbol {\theta}} _k}}$ with the global parameters $\boldsymbol{\theta}_G$. The local models then perform supervised learning to minimize the loss $L_{\rm s}(\boldsymbol{\theta}_k)$ on their specific subdatasets $\mathcal{D}^k$. Subsequently, $G$ is obtained by aggregating the parameters learned locally, ${\boldsymbol{\theta}}_G \leftarrow \frac{N^k}{N} \sum_{k=1}^K \boldsymbol{\theta}_k$, and distributes the updated parameters back to the local models for the next round. This iterative process continues until the final round $T$.

    SSFL can be employed to address the issue of scarce labels among clients in FL by utilizing unlabeled data. In this scenario, each dataset $\mathcal{D} = \{\boldsymbol{x}_i, {\boldsymbol y}_i\}_{i = 1}^N$ is divided into a labeled subset $\mathcal{S} = \{\boldsymbol{x}_i, {\boldsymbol y}_i\}_{i = 1}^S$ and an unlabeled subset $\mathcal{U} = \{\boldsymbol{u}_i\}_{i = 1}^U$. Within the federated framework, a series of local models manage these subsets, with each client possessing a portion of both labeled $\mathcal{S}^k = \{\boldsymbol{x}_i^k, {\boldsymbol y}_i^k\}_{i = 1}^{S^k}$ and unlabeled data $\mathcal{U}^k = \{\boldsymbol{u}_i^k\}_{i = 1}^{U^k}$, distributed privately across $K$ clients, where ${S^k}$ and ${U^k}$ represent the number of labeled and unlabeled samples of the $k$-th client, respectively. 
    
    In Section \ref{3}, we will introduce our proposed SSFL-DCSL framework in detail to address the issues of data distribution discrepancies and insufficient labels. Through prototype-based communication, SSFL-DCSL enables the local model to align its prototype with other local models while minimizing the total loss ${L_{\rm {total}}}$ (Eq. ($\ref{totalloss}$)) of all client-side local learning tasks.

\section{Proposed Method}\label{3}

    In this section, we first introduce the overall framework of the proposed SSFL-DCSL. Then, the key components of the SSFL-DCSL are described in detail.

\subsection{Overall SSFL-DCSL Framework}
    The overall SSFL-DCSL framework consists of three main stages: training, fine-tuning, and testing. The diagram of the training stage is illustrated in Fig. \ref{fig1}.

   During the training stage of the SSFL-DCSL framework, the following steps are executed for both labeled and unlabeled data in each round: (1) Clients concurrently train their local models. (2) Clients send updates of their local prototypes ${ {{\boldsymbol P}^1}, \cdots ,{{\boldsymbol P}^{K}}} $ to the server.
    (3) The server collects these local prototypes and performs weighted aggregation. (4) The server redistributes the updated global prototypes ${{\boldsymbol P}}$ to the clients.
    Steps (1) to (4) are repeated until the model converges or the maximum number of iterations is reached.
    
    In particular, in Step (1), we design a TLAW function to address low confidence in pseudo labels. Additionally, we introduce a dual contrastive loss, which addresses the issues of insufficient data labels and inconsistent data distributions by modeling feature distribution differences from two perspectives, namely local contrastive loss (LCL) and global contrastive loss (GCL). In Step (3), to alleviate the instability of global prototype updates, we perform momentum updates on prototypes between rounds.
    
    During the fine-tuning stage, each client fine-tunes the model using its locally labeled data. During the testing stage, test samples are input into the model to predict fault types.
    
    The details of each component in the training stage are described in the following subsections.

\subsection{ Local Training}
    The local training on the clients in Step (1) adopts the paradigm called pseudo-labeling\cite{lee2013pseudo}, which applies training to both labeled and unlabeled data for each user. To counteract the problem of low-quality pseudo-labels that adversely affect training performance, a TLAW function is proposed. Meanwhile, a dual contrastive loss is designed to address the challenges of insufficient data labels and variations in data distribution.

\subsubsection{Semi-Supervised Learning}
    For a training sample $\boldsymbol{x}$, and its corresponding label $\boldsymbol{y}$, we define ${\boldsymbol{p}_{\boldsymbol{\theta}} }(\boldsymbol{y}|\boldsymbol{x}) \in {\mathbb{R}^C}$ as the prediction of the model. $\mathcal{H}(\boldsymbol{a},\boldsymbol{b})$ represents the cross-entropy loss, where $\boldsymbol{a}$ represents the true probability distribution of the labels and $\boldsymbol{b}$ represents the predicted probability distribution of the model.  

    The loss function for SSL consists of two cross-entropy loss terms: a supervised loss ${L_{\rm s}}$ applied to labeled data and an unsupervised loss ${L_{\rm u}}$ applied to unlabeled data calculated based on the pseudo-label (which is introduced in Section \ref{lu}).  Specifically, ${L_{\rm s}}$ is the standard cross-entropy loss, given by
    \begin{equation}
    	{L_{\text{s}}} = \frac{1}{E}\sum\limits_{i = 1}^E {{\mathcal{H}}({{\boldsymbol y}_i},{p_{{{\boldsymbol{\theta}}_k}}}({\boldsymbol y}|{{\boldsymbol x}_i}))} ,
    	\label{Ls}
    \end{equation}
    where $E$ represents the number of labeled samples.

    
    The total loss function of standard SSL can be expressed as ${L_{\rm s}} + \eta (t){L_{\rm u}}$, where $\eta (t)$ is a dynamic coefficient used to balance $L_{\rm s}$ and $L_{\rm u}$ at the $t$-th round. The value of $\eta (t)$ is determined based on the settings in \cite{lee2013pseudo}, with ${\eta _f} = 3$, ${T_1} = 0.3*T$, ${T_2} = 0.7*T$, and $T$ is global rounds:
    \begin{equation}
    \eta (t) = \begin{cases}
    	0, &  {\rm{t}} < T1, \\
    	\frac{{{\rm{t}} - {T_1}}}{{{T_2} - {T_1}}}{\eta  _f}, &  {T_1} \le t < {T_2},\\
    	{\eta _f}, &  {T_2} \le t.
    \end{cases}
    \end{equation}


\subsubsection{Truncated Laplace-based Adaptive Sample Weighting Function}\label{lu}
    During SSL training process, indiscriminately using low-confidence pseudo labels can exacerbate the bias in predictions. To address this, we introduce a strategy that employs a truncated Laplace distribution to weight pseudo-labels based on their confidence levels, thus reducing bias.

    The calculation of unsupervised loss $L_{\rm u}$ requires pseudo-labels, and existing methods often use a threshold to filter out low-confidence pseudo-labels during training \cite{FixMatch}. In this paper, we tackle the issue of inaccurate pseudo-labels from a sample weighting perspective. Specifically, we define the unsupervised loss ${L_{\rm u}}$ as the weighted cross-entropy, given as
    \begin{equation}
        {L_{\rm u}}  = \frac{1}{O}\sum\limits_{i = 1}^O {\lambda (\boldsymbol p_i)\mathcal{H}({{\hat {\boldsymbol p}}_i},{\boldsymbol p}_{{\boldsymbol{\theta} _k}}({\boldsymbol y}|\mathcal{A}({{\boldsymbol{u}}_i})))} ,
        \label{LU}
    \end{equation}
	where $\boldsymbol p$ is the abbreviation of ${\boldsymbol p}_{\boldsymbol{\theta} _{k}}({\boldsymbol y}|\alpha ({{\boldsymbol{u}}_i}))$, and ${{\hat {\boldsymbol p}}_i} = \arg \max ({{\boldsymbol p}_i})$ is served as the one-hot pseudo label. $\lambda (\boldsymbol p)$ is a sample weighting function with a range of $ [{0,{\lambda _{\max }}}]$, and $O$ represents the number of pseudo labels. $\alpha ({{\boldsymbol{u}}_i})$  
    and $\mathcal{A}({{\boldsymbol{u}}_i})$ are the weak and strong augmentations of unlabeled data, respectively. 
 
    It is noted that data augmentation for unlabeled data is an important step of the pseudo-label method \cite{lee2013pseudo}. To enhance the robustness of representation learning, we employ two different augmentations, such that one augmentation is weak and the other is strong, following the methods in \cite{ijcai2021aug}. In particular, weak augmentation $ \alpha (\boldsymbol{u}_i)$ includes jitter and scaling to introduce random variations and amplify the magnitude of the signal. Strong augmentation $ \mathcal{A}(\boldsymbol{u}_i)$ combines permutation and random jitter, which divides the signal into random segments up to size $M$ and shuffles them, then adds random jitter to the rearranged signal.
	
	We assume that the confidence of the model outputs $\max(\boldsymbol{p})$  follows a dynamic truncated Laplace distribution with mean $\mu_t$ and scale parameter $b_t$ at the $t$-th iteration and ${b_t} = \sqrt {\frac{1}{2}\sigma _t^2} $, where $\sigma_t$ is the standard deviation. We propose using the deviation of $\max(\boldsymbol{p})$ and the mean $\mu_t$ of the Laplace distribution as a measure of the accuracy of model prediction, where samples with higher confidence are less prone to errors compared to those with lower confidence. Finally, $\lambda ({\rm{\boldsymbol{p}}})$ can be expressed as
	\begin{equation}
		\lambda(\boldsymbol{p}) = \begin{cases}
			\lambda_{\max} \cdot 2b_t \cdot \phi(\max({\boldsymbol p}; \mu_t, b_t)), & \text{if } \max(\boldsymbol{p}) < \mu_t, \\
			\lambda_{\max}, & \text{otherwise},
		\end{cases}
	\label{eq6}
	\end{equation}
	which is a truncated Laplace distribution regarding $\max(\boldsymbol{p})$ with a range of $\lambda ({\rm{\boldsymbol{p}}}) \in \left[ {0,{\lambda _{\max }}} \right]$. In (\ref{eq6}), $\phi (\boldsymbol x;\mu ,b) = \frac{1}{{2b}}\exp ( - \frac{{|\boldsymbol x - \mu |}}{b})$ and ${\lambda _{\max }}$ are constants, representing the weight coefficient.
	
	However, the parameters $\mu_t$ and $\sigma_t$ are both unknown, and we can estimate them from historical predictions of the model. At the $t$-th iteration, the batch mean ${{\hat \mu }_{\rm{b}}}$ and variance $\hat \sigma _b^2$ are calculated as
	 \begin{equation}
	 	\begin{array}{l}
	 		{{\hat \mu }_{\rm{b}}} = \frac{1}{{{B_U}}}\sum\nolimits_{i = 1}^{{B_U}} {\max ({{\boldsymbol{p}}_i})} ,\\
	 		\hat \sigma _b^2 = \frac{1}{{{B_U}}}\sum\nolimits_{i = 1}^{{B_U}} {(\max ({{\boldsymbol{p}}_i})}  - {{\hat \mu }_{\rm{b}}}{)^2},
	 	\end{array}\\
	 \end{equation}
    where $B_U$ is the number of unlabeled samples in a batch.
 
 	To avoid bias caused by single-batch estimates, we use exponential moving average (EMA) with momentum $m$ for more stable estimation:
 	\begin{equation}
 	\begin{array}{l}
 		{{\hat \mu }_{\rm{t}}} = m{{\hat \mu }_{{\rm{t - 1}}}} + (1 - m){{\hat \mu }_{\rm{b}}},\\
 		\hat \sigma _t^2 = m\hat \sigma _{t - 1}^2 + (1 - m)\frac{{{B_U}}}{{{B_U} - 1}}\hat \sigma _b^2,
 	\end{array}
 	\end{equation}
     where we use unbiased variance for EMA, and initialize ${{\hat \mu }_0}$ as $\frac{1}{C}$ and $\hat \sigma _0^2$ as 1.0. The estimated mean ${{\hat \mu }_{\rm{t}}}$ and variance $\hat \sigma _t^2$ are fed back into (\ref{eq6}) to calculate the weights of the samples.
	
	To demonstrate the effectiveness of TLAW, we analyze it from two perspectives: the quantity and quality of the pseudo-labels. 

    \begin{definition}
     Define the quantity of pseudo-labels, $f(\boldsymbol{p})$, as the average sample weight of unlabeled data when calculating the weighted unsupervised loss, given by
	\begin{equation}
		f(\boldsymbol{p}) = \sum\nolimits_i^U {\frac{{\lambda ({\boldsymbol{p}_i})}}{U}}  = {\mathbb{E}_U}[\lambda ({\boldsymbol{p}_i})],
	\end{equation}
	where each unlabeled data is sampled evenly in ${\mathcal{U}}$.
    \end{definition}
    
    \begin{definition}
    Define the quality of pseudo-labels, $g(\boldsymbol{p})$, as the percentage of correct pseudo-labels, assuming the ground truth ${{\boldsymbol y}^{\rm u}}$ is known, given by

    \begin{equation}
    g(\boldsymbol{p}) = 
    {E_{\bar \lambda (\boldsymbol{p})}}[\mathds{1}(\arg \max (\boldsymbol{p}) = {{\boldsymbol y}^{\rm u}})] ,
    \end{equation}
	where $\bar \lambda (\boldsymbol{p}) = \lambda (\boldsymbol{p})/\sum {\lambda (\boldsymbol{p})} $. $\mathds{1}(i=j) $ is an indicating function, if $i=j$, $\mathds{1}=1$; if $i \ne j$, $\mathds{1}=0$.
 
    \end{definition}

     \begin{theorem}\label{thm:main}
    The sample weighting function based on the Laplace distribution guarantees at least the quantity of $\frac{{{\lambda _{\max }}}}{2}$ and at least the quality of $\sum\nolimits_j^{\hat U} {\frac{{\mathds{1}({{\hat {\boldsymbol p}}_i} = {\boldsymbol y}_j^\mathrm{u})}}{{2\hat U}}} $. where $\hat U = \sum\nolimits_i^U {\mathds{1}(\max ({\boldsymbol{p}_i})}  \ge {\mu _t})$. 
    \end{theorem}
    
   The above theorem assigns lower weights to low-quality pseudo labels based on confidence deviation, allowing the model to focus primarily on high-quality pseudo-labels. This approach enables the model to learn more accurate and useful information, thereby improving overall performance. However, pseudo labels of poor quality and good quality are given the same weight in \cite{lee2013pseudo}. This can lead the model to learn incorrect information, thereby affecting its performance.  The proof of \textbf{Theorem} \ref{thm:main} can be found in Appendix \ref{proof}.

\subsubsection{Dual Contrastive Loss}
    To tackle insufficient data labels and disparate data distributions, we further introduce a dual contrastive loss in local training. Users utilize local unlabeled data to learn meaningful representations by assessing sample similarity and dissimilarity, reducing dependence on labeled data. Furthermore, comparing each user's local prototypes against global prototypes helps to narrow the distance between them.
	
	The DL model comprises two components: (1) the representation layers, which use embedding functions to transform raw inputs into an embedded space, and (2) the decision layers, which handle classification tasks. Specifically, the embedding function of the $k$-th client is ${{f}_k}({{\boldsymbol{\vartheta}} _k})$ parameterized by ${{\boldsymbol{\vartheta}} _k}$. We denote ${{\boldsymbol c}_k} = {f_k}({{\boldsymbol{\vartheta}} _k};{\boldsymbol{x}})$ as the embedding of $\boldsymbol{x}$.
	For a given classification task, the predictions for $\boldsymbol{x}$ can be generated by a function ${g_k}({{\boldsymbol{\varphi}} _k})$ parameterized by ${{\boldsymbol{\varphi}} _k}$. Therefore, the model function can be written as ${F_k}{\rm{ = }}{g_k}({{\boldsymbol{\varphi}} _k}) \circ {f_k}({{\boldsymbol{\vartheta}} _k})$, which we abbreviate as ${\boldsymbol{\theta} _k}$ to denote $({{\boldsymbol{\vartheta}} _k},{{\boldsymbol{\varphi}} _k})$.
	
    CL aims to maximize the similarity between different views of the same sample while minimizing its similarity to other samples. As mentioned in Section \ref{lu}, the unlabeled data are augmented strongly and weakly, denoted as ${{\boldsymbol{z}_i}^w} = \alpha (\boldsymbol{u}_i)$ and ${{\boldsymbol{z}_i}^s} = \mathcal{A}(\boldsymbol{u}_i)$ respectively. Hence, the features extracted by the representation layers are denoted as ${{\boldsymbol{c}_i}^w} = f({{\boldsymbol{z}_i}^w})$ and ${{\boldsymbol{c}_i}^s} = f({{\boldsymbol{z}_i}^s})$. Naive CL treats all samples equally and uses a fixed temperature, which will amplify noises from unreliable pseudo-labels and fail to adapt to batch-wise confidence fluctuations. We therefore introduce a pseudo-label-guided, confidence-weighted LCL that (1) selects positive/negative pairs (SPNP) according to pseudo-labels, and (2) adopts a dynamic temperature (DT) to stabilize training. The dynamic temperature is defined as $\tau_i = \tau\bigl(1 + \alpha\,\sigma_t\bigr)$, where \(\tau\) denotes the base temperature and \(\alpha\) is the temperature scaling factor. To learn the differences between sample features and enhance the richness and discriminative power of the representations, the LCL is utilized, given as
    \begin{equation}
    L_{\rm {LC}}
    = -\frac{1}{B}
      \sum_{i=1}^{B}
      \log\frac{N_i}{D_i}\,,
      \label{llc}
    \end{equation}
    where
    \begin{equation}
        \begin{aligned}
        N_i &= \sum_{j\in\mathcal{P}_i}\exp\!\bigl(\mathrm{sim}(\boldsymbol c_i^w,\,\boldsymbol c_j^s)/\tau_i\bigr)\,,\\
        D_i &= N_i 
             + \sum_{k\in\mathcal{N}_i}\exp\!\bigl(\mathrm{sim}(c_i^w,\,c_k^s)/\tau_i\bigr)\,,
        \end{aligned}
    \end{equation}
    with
    \begin{equation}
        \mathcal{P}_i = \bigl\{\,j \mid \hat y(\boldsymbol{c}_j^s)=\hat y(\boldsymbol{c}_i^w)\bigr\},
        \quad
        \mathcal{N}_i = \bigl\{\,k \mid \hat y(\boldsymbol{c}_k^s)\neq\hat y(\boldsymbol{c}_i^w)\bigr\}.
    \end{equation}
 and $B$ is the batch size.
	
	For the $k$-th user's local prototypes ${\boldsymbol P}_j^{k}$ and global prototypes ${{\tilde {\boldsymbol P}}_{j}}$ (see Section \ref{prototypesBasedAggregation} for prototypes computation), the GCL is designed to minimize the disparity between local and global prototypes. It enforces consistency between different augmented views of the same sample, enabling the network to learn robust, discriminative representations without relying on labels.  By leveraging inherent structure in unlabeled data through self-supervised local training, this approach compensates for the scarcity of labeled samples and mitigates feature drift caused by heterogeneous local data distributions. The loss function is given by
	\begin{equation}
		{L_{\rm {GC}}} =  - \sum\nolimits_{i = 1}^C 	{\log \frac{{\exp ({\rm{sim}}({{\boldsymbol P}_{i}},{{\tilde {\boldsymbol P}}_{i}})/{\tau _i})}}{{\sum\nolimits_{j = 1}^C {{\mathds{1}_{[j \ne i]}}\exp ({\rm{sim}}({{\boldsymbol P}_{i}},{{\tilde {\boldsymbol P}}_{j}})/{\tau _j})} }}} .
        \label{lgc}
	\end{equation}

     GCL aligns local feature representations with globally aggregated prototypes across multiple clients, addressing data heterogeneity and enriching the learned feature space with inter-client information. This global alignment enhances the model's robustness by leveraging collaborative knowledge from different clients, thereby compensating effectively for limited labeled data. By jointly optimizing LCL and GCL, the method integrates self-supervised local training and global collaborative learning, significantly improving the model's generalization capabilities.

     In the SSFL-DCSL framework, local models aim to align their prototypes with the global prototypes while minimizing a composite loss comprising supervised loss ${L_{\rm s}}$, unsupervised loss ${L_{\rm u}}$, LCL ${L_{LC}}$ and GCL ${L_{GC}}$.  The overall loss function is given by
	\begin{equation}
		{L_{\rm {total}}} = {L_{\rm s}} + \eta (t){L_{\rm u}} + {L_{LC}} + \iota (t) {L_{GC}},
        \label{totalloss}
	\end{equation}
	where $\iota (t)$ is a weighting coefficient. Define $E$ as the number of labeled samples within a batch, we have $\iota (t) = E/B$.

    The objective of the $k$-th client can be expressed as
    \begin{equation}
        \arg \mathop {\min }\limits_{{\boldsymbol{\theta }_k} = ({\boldsymbol{\vartheta} _k},{\boldsymbol{\varphi} _k})}  L_{\rm s}^k({\boldsymbol{\theta} _k}) + \eta (t) \cdot L_{\rm u}^k({\boldsymbol{\theta} _k}) + L_{\rm {LC}}^k({\boldsymbol{\vartheta} _k}) + \iota (t) \cdot L_{\rm {GC}}^k({\boldsymbol{\vartheta} _k})
    \end{equation}
    where \( \boldsymbol{\theta}_k \) denotes the local model parameters of client \( k \), including the feature extractor \( \boldsymbol{\vartheta}_k \) and classifier \( \boldsymbol{\varphi}_k \). \( L_{\rm s}^k \) is the supervised loss on labeled data (Eq.~(\ref{Ls})), and \( L_{\rm u}^k \) (Eq.~(\ref{LU})) is the confidence-weighted pseudo-label loss on unlabeled data. \( L_{\text{LC}}^k \) (Eq.~(\ref{llc})) is the LCL for consistency across augmented views, while \( L_{\text{GC}}^k \) (Eq.~(\ref{lgc})) is GCL aligning local prototypes \( \boldsymbol {P}_j^k \) with global prototypes \( \tilde{\boldsymbol {P}}_j \). \( \eta(t) \) and \( \iota(t) \) are dynamic weights for the unsupervised and global contrastive losses, adjusted by communication round.

    Once the local training objectives are established, coordinated aggregation across clients becomes critical for attaining global feature alignment. The following section details our prototypes-based aggregation strategy.

\subsection{Prototypes-Based Aggregation}\label{prototypesBasedAggregation}
	Due to the uneven distribution of user samples and to mitigate the impact of data skewness, and to better balance the performance of the model across different classes, we adopt a PTA approach. It aggregates local prototypes into a global prototype space, mitigating performance issues from sample heterogeneity.  Specifically, users compute and upload the average high-dimensional feature vectors of each category as prototypes for aggregation in each iteration. Meanwhile, by generating and sharing abstract representations of data locally instead of the actual data, privacy is effectively protected and the risk of data leakage is reduced.
	
    Define ${{\boldsymbol P}_{{j}}}$ as the prototypes for $j$-th class. For the $k$-th client, the prototypes are the average value of the embedding vectors of samples in class $j$, given by
    \begin{equation}
        {\boldsymbol P}_j^{k} = \frac{1}{{|\mathcal{D}_j^k|}}\sum\limits_{({\boldsymbol x},{\boldsymbol y}) \in {\mathcal{D}_j^k}} {{f_k}(} {{\boldsymbol {\vartheta}} _k};{\boldsymbol x}),
        \label{Localprototypes}
    \end{equation}
	where $\mathcal{D}_j^k$ represents a subset of the local dataset $\mathcal{D}^k$ containing the $j$-th class, including labeled data of $j$-th class and unlabeled data with the pseudo label $j$.
	
	Given a class $j$, the server receives prototypes from a group of users with class $j$. After the aggregation operation of the prototypes, a global prototypes ${{\tilde {\boldsymbol P}}_{j}}$ of class $j$ is generated. The global prototypes of the $j$-th class is represented as
	\begin{equation}
		{{\tilde {\boldsymbol P}}_{j}} = \frac{1}{{|{\mathcal{N}_j}|}}\sum\limits_{k \in {\mathcal{N}_j}} {\frac{{|\mathcal{D}_j^k|}}{{{N_j}}}{\boldsymbol P}_j^{k}} ,	
        \label{globalprototypes}
	\end{equation}
	where ${N_j}$ indicates the total number of samples of the $j$-th class of all users, and ${\mathcal{N}_j}$ indicates the set of users that have class $j$.

	Variations in feature distributions between pseudo-labeled and labeled samples, along with differences in user data distribution, result in inconsistent prototypes. Directly updating the global prototypes can cause significant disparities between iterations, degrading training performance. Moreover, incorrect pseudo-labels further undermine the robustness and stability of prototypes updates.

	To stabilize prototypes updates, we propose a momentum-based method that uses an exponentially weighted moving average of gradients to update model weights. We apply this approach to the global prototypes at each iteration, thereby retaining aspects of previously updated prototypes. In the $t$-th iteration, the global prototypes ${\tilde {\boldsymbol P}}$, derived from the current user's local prototypes, given by
	\begin{equation}
	{{\tilde {\boldsymbol P}}^{\rm{t}}} = [{{\tilde {\boldsymbol P}}_1^{t}},{{\tilde {\boldsymbol P}}_2^{t}}, \ldots ,{{\tilde {\boldsymbol P}}_C^{t}}],
	\end{equation}

	\begin{equation}
    		{{\boldsymbol P}^{t + 1}} = \kappa {{\boldsymbol P}^t} + (1 - \kappa ){{\tilde {\boldsymbol P}}^{{{t + 1}}}}.
            \label{ema}
    	\end{equation}

	Subsequently, in the $t+1$ iteration, the updated global prototypes ${{\boldsymbol P}^{t + 1}}$ combine the current and the previous prototypes to reduce the impact of data skewness and make the update of the global prototypes more stable:
    
    The detailed algorithm of SSFL-DCSL is summarized in \textbf{Algorithm}. \ref{algorithm}. The client performs local training and updates the local prototypes to the server. The server then calculates the global prototypes and sends them to each client. This process is repeated until the model converges.

\section{Experiments}\label{4}

    To evaluate the proposed SSFL-DCSL framework, we conduct experiments on three publicly available datasets.  The experimental results are discussed and further compared with the existing methods.

    \begin{algorithm}[]
      
		\SetKwData{Left}{left}\SetKwData{This}{this}\SetKwData{Up}{up}
		\SetKwFunction{Union}{Union}\SetKwFunction{FindCompress}{FindCompress}
		\SetKwInOut{Input}{Input}\SetKwInOut{Output}{output}
		
		\caption{The proposed SSFL-DCSL algorithm}\label{algorithm}
		\Input{${\mathcal{D}^k}$, ${\boldsymbol{\theta} _k}$, ${k} = 1, \ldots ,K$ }
		
		\For{each round $t=1,2,\ldots $}{
                \For{each client $k$ \textbf{ in parallel}}{
    		\For{ $({\boldsymbol{x}_i},{{\boldsymbol y}_i}) \in \mathcal{S}^k,{{\boldsymbol{u}}_i} \in \mathcal{U}^k$}{
                     Compute local prototypes $\boldsymbol P^{k}$ by Eq. (\ref{Localprototypes}).\par
                     Compute the loss ${L_{total}}$ by Eq. (\ref{totalloss}) using local prototypes.\par
                     Update local model ${{{\boldsymbol {\theta}} _k}}$  according to the loss.
                    }
                
                }
                
			Update global prototypes ${\boldsymbol P}$ by Eq. (\ref{globalprototypes})-(\ref{ema}).\par
                Update local prototypes set $\{{\boldsymbol P}^k\}$ with prototypes in $\{ {{\tilde {\boldsymbol P}}_{j}}\} $
		}		
    \end{algorithm}

\subsection{Dataset and Experimental Setup}

\subsubsection{Benchmark datasets}

   The experimental confirmation involved the use of three openly accessible datasets: the dataset from Purdue University (PU), the dataset from Mechanical Fault Prognosis Technology (MFPT), the dataset from Case Western Reserve University (CWRU), and the chemical plant dataset (CP) collected from 4 pumps in the chemical factory.

    \textbf{PU:}
    The PU dataset includes data collected from a test setup featuring an electric motor, a torque measurement shaft, a test module, a flywheel, and a load motor. It records vibration signals at a sampling rate of 64,000 Hz. The dataset includes four operating conditions and three states: healthy, inner race fault, and outer race fault. The sample length is 2048, and the number of samples is 8348. The data partitioning method is the same as the PU dataset.

    \textbf{MFPT:}
    The MFPT dataset features three operating conditions: normal, outer race fault, and inner race fault. Under normal conditions, the dataset records a load of 270 pounds, an input shaft speed of 25 Hz, and a sampling frequency of 97,656 Hz for six seconds. Both fault conditions maintain a 25 Hz shaft speed with a sampling frequency of 48,828 Hz. The sample length is 2048, and the number of samples is 1852. The data are processed in the same way as the PU dataset. 

    \textbf{CWRU:}
    The CWRU dataset includes a healthy state and three fault types: inner race, outer race, and ball faults. Data were captured using accelerometers at both the drive and fan ends of the bearings, with sampling frequencies of 12,000 or 48,000 Hz. The dataset covers four operating conditions with bearing loads ranging from 0 to 3 horsepower. The sample length is 2048, and the number of samples is 17,444. The data are processed in the same way as the other datasets. 

\begin{figure}[]
    \centering
    \begin{minipage}{0.33\linewidth} 
        \centering
        \includegraphics[scale=0.11]{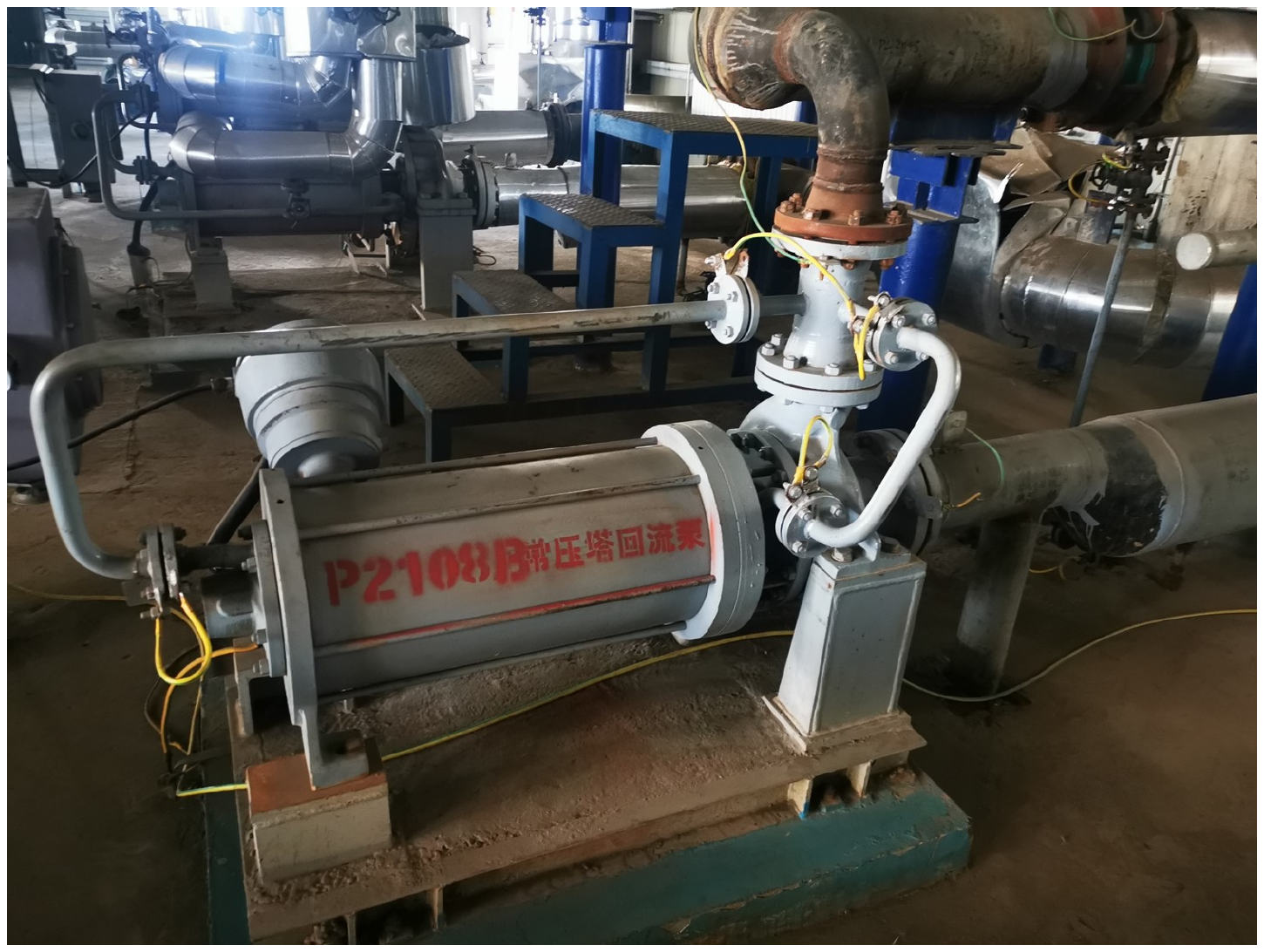}
        \vspace{-6mm}
        \caption*{\white{bbb}(a)}
    \end{minipage}%
    \hfill 
    \begin{minipage}{0.33\linewidth} 
        \centering
        \includegraphics[scale=0.11]{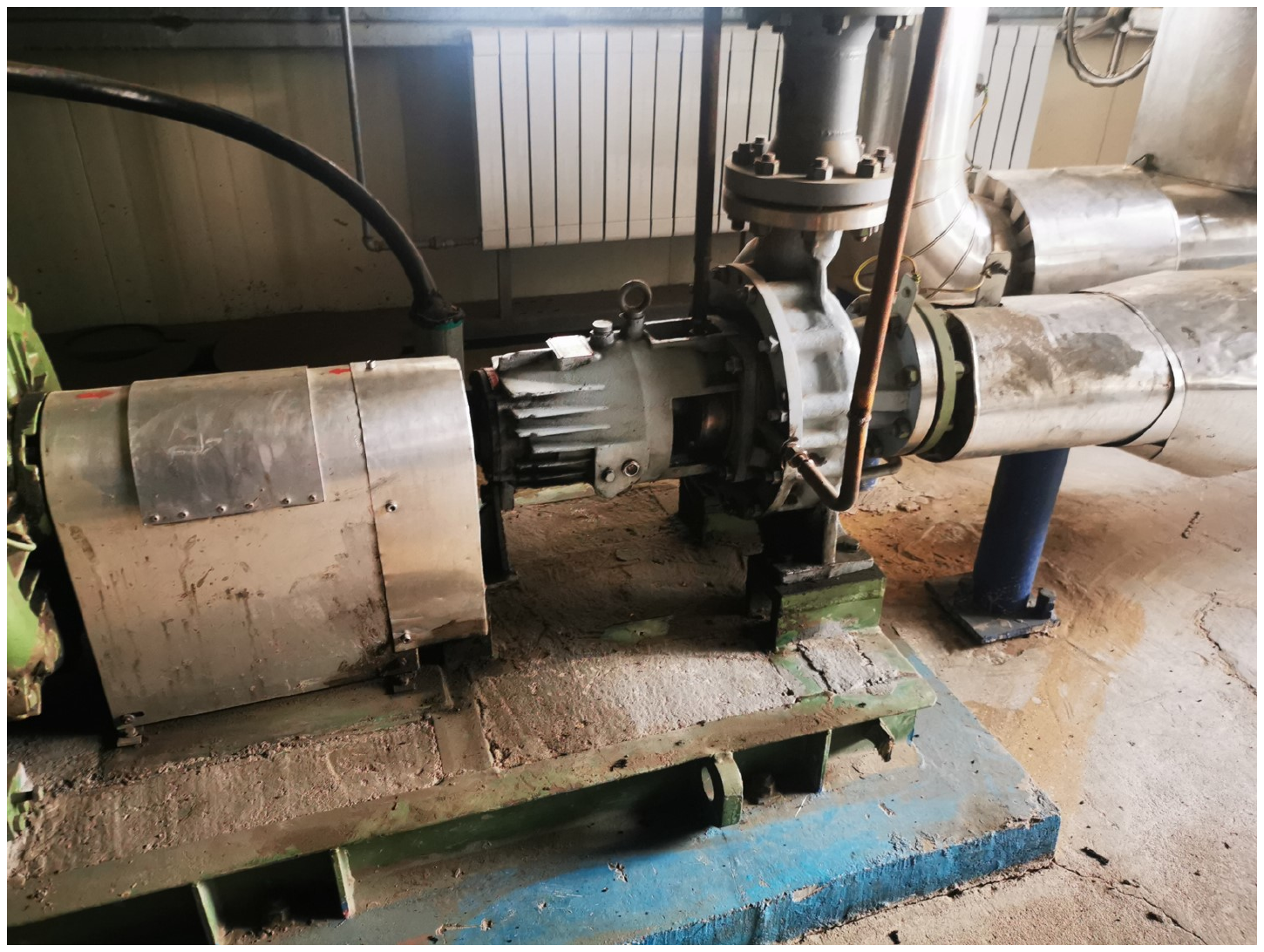}
        \vspace{-6mm}
        \caption*{\white{bbb}(b)}
        
    \end{minipage}%
    \hfill 
    \begin{minipage}{0.33\linewidth}
        \centering
        \includegraphics[scale=0.11]{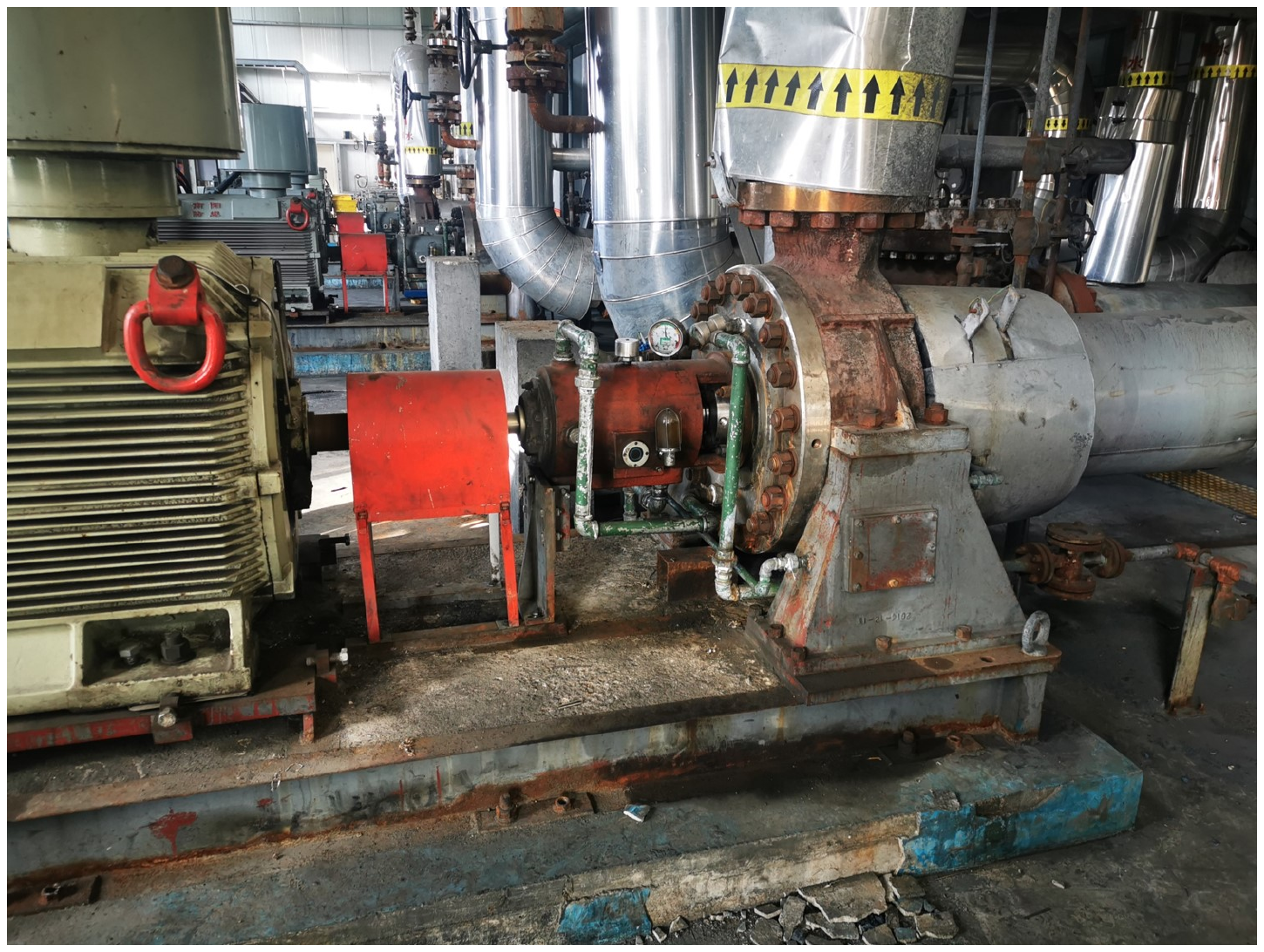}
        \vspace{-6mm}
        \caption*{\white{bbb}(c)}
    \end{minipage}
    
    \caption{The machines in our partner chemical factory. (a) Reflux Pump. (b) Feeding Pump. (c) Circulating Pump.}
    \label{pumps}
\end{figure}

\begin{figure}[]
    \centering
    \begin{minipage}{0.48\linewidth} 
        \centering
        \includegraphics[scale=0.03]{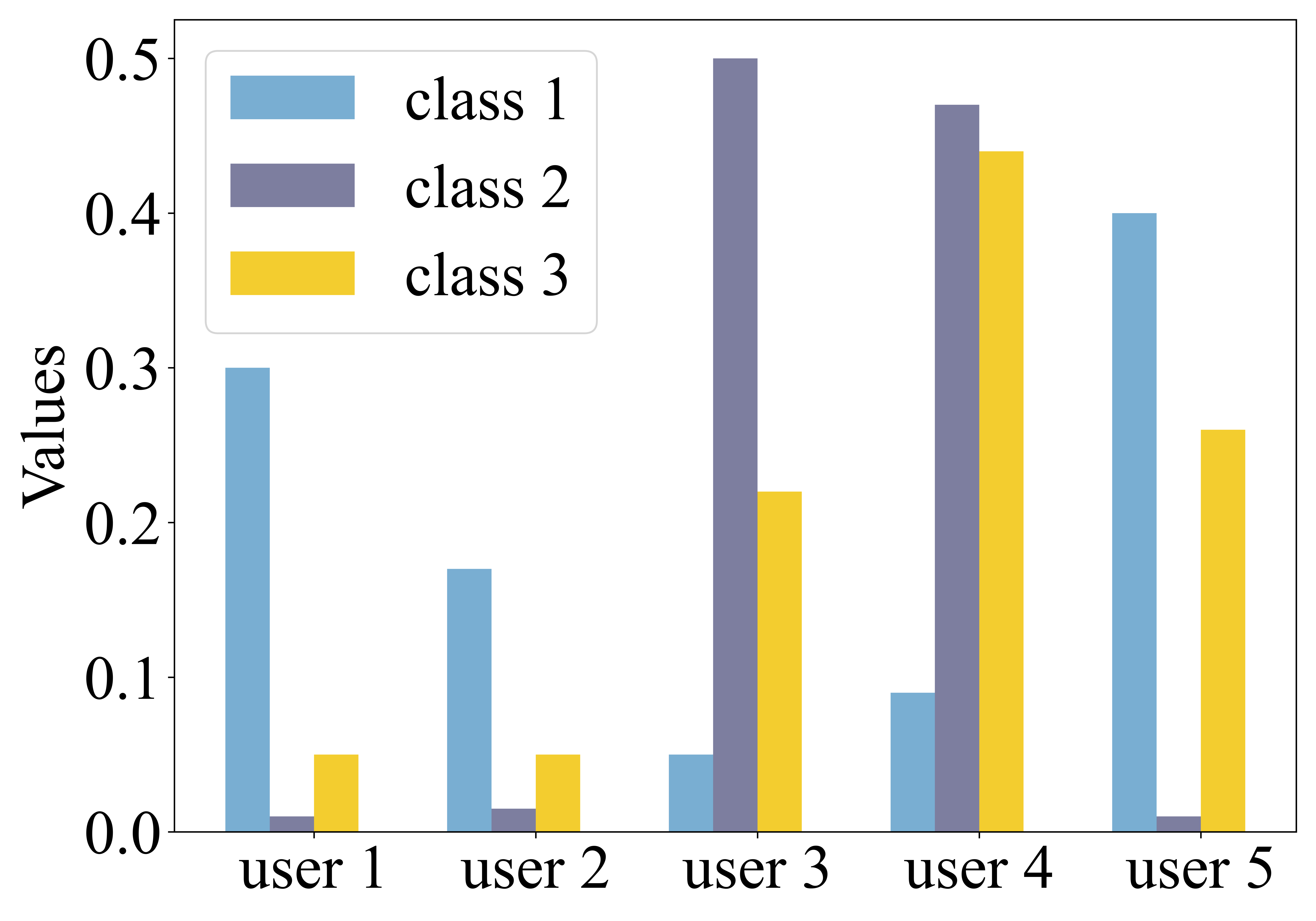}
        \caption*{\white{b}(a) PU}
    \end{minipage}%
    \hfill 
    \begin{minipage}{0.48\linewidth}
        \centering
        \includegraphics[scale=0.03]{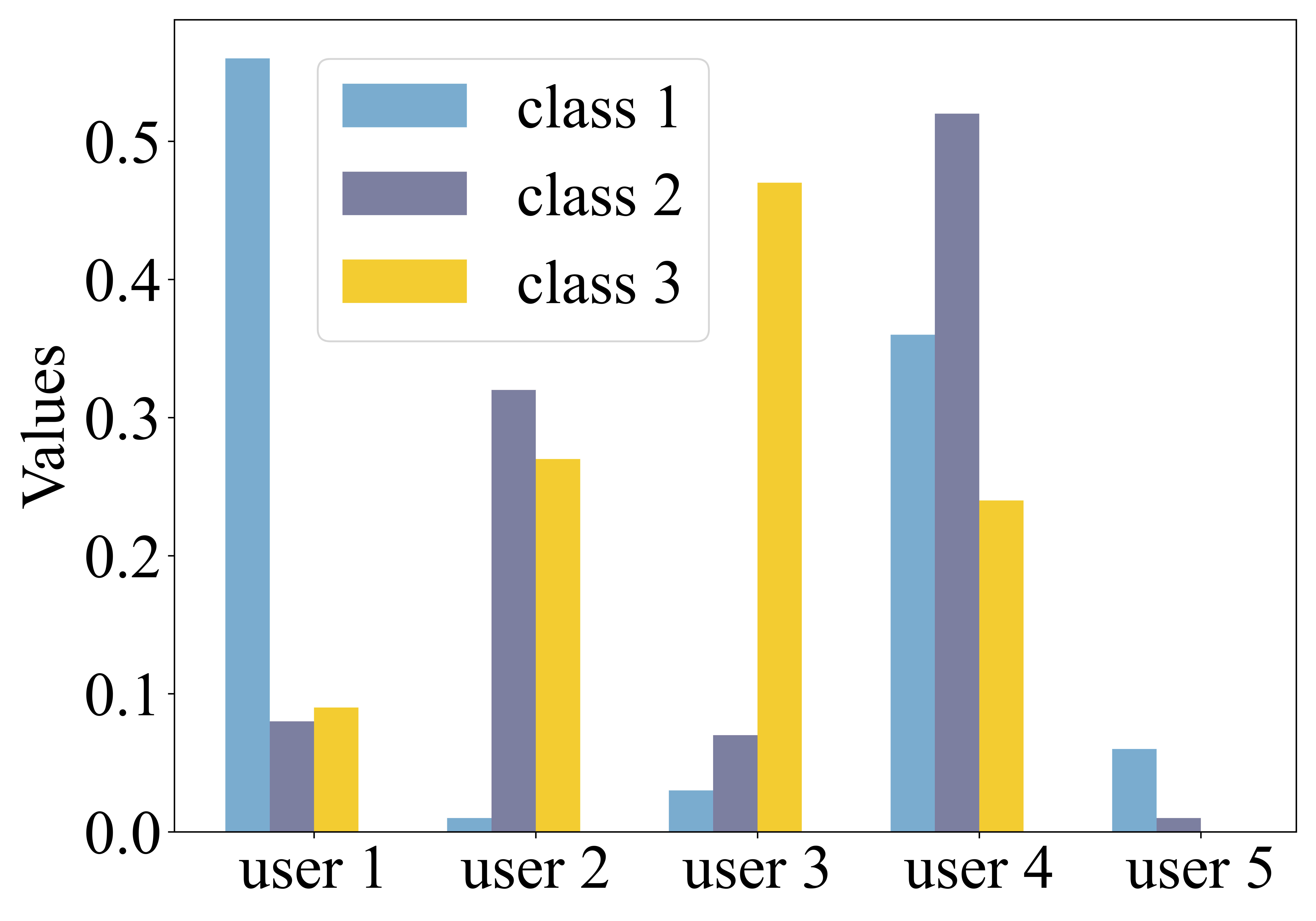}
        \caption*{\white{b}(b) MFPT}
    \end{minipage}

    \begin{minipage}{0.48\linewidth} 
        \centering
        \includegraphics[scale=0.03]{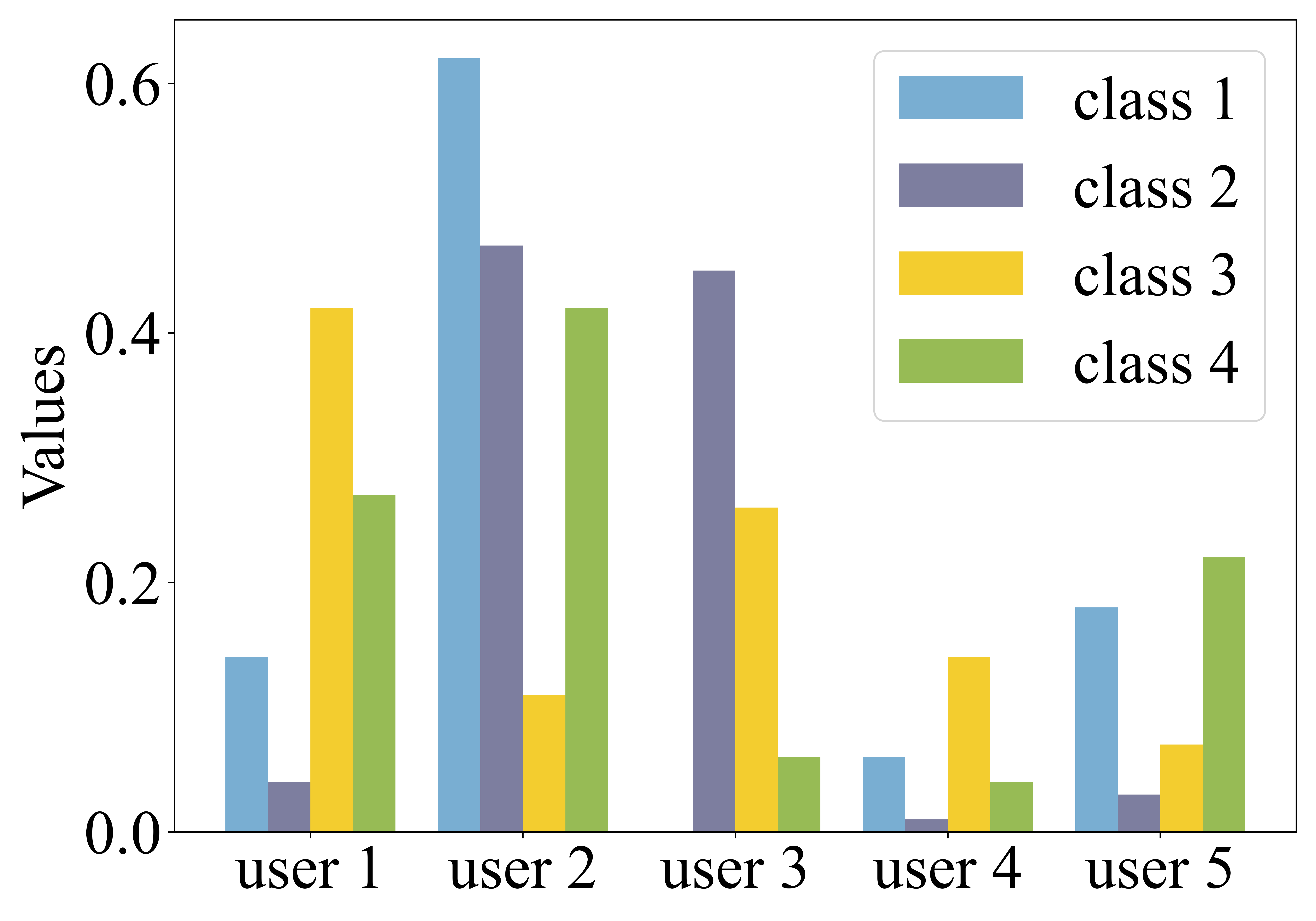}
        \caption*{\white{b}(c) CWRU}
    \end{minipage}%
    \hfill 
    \begin{minipage}{0.48\linewidth}
        \centering
        \includegraphics[scale=0.03]{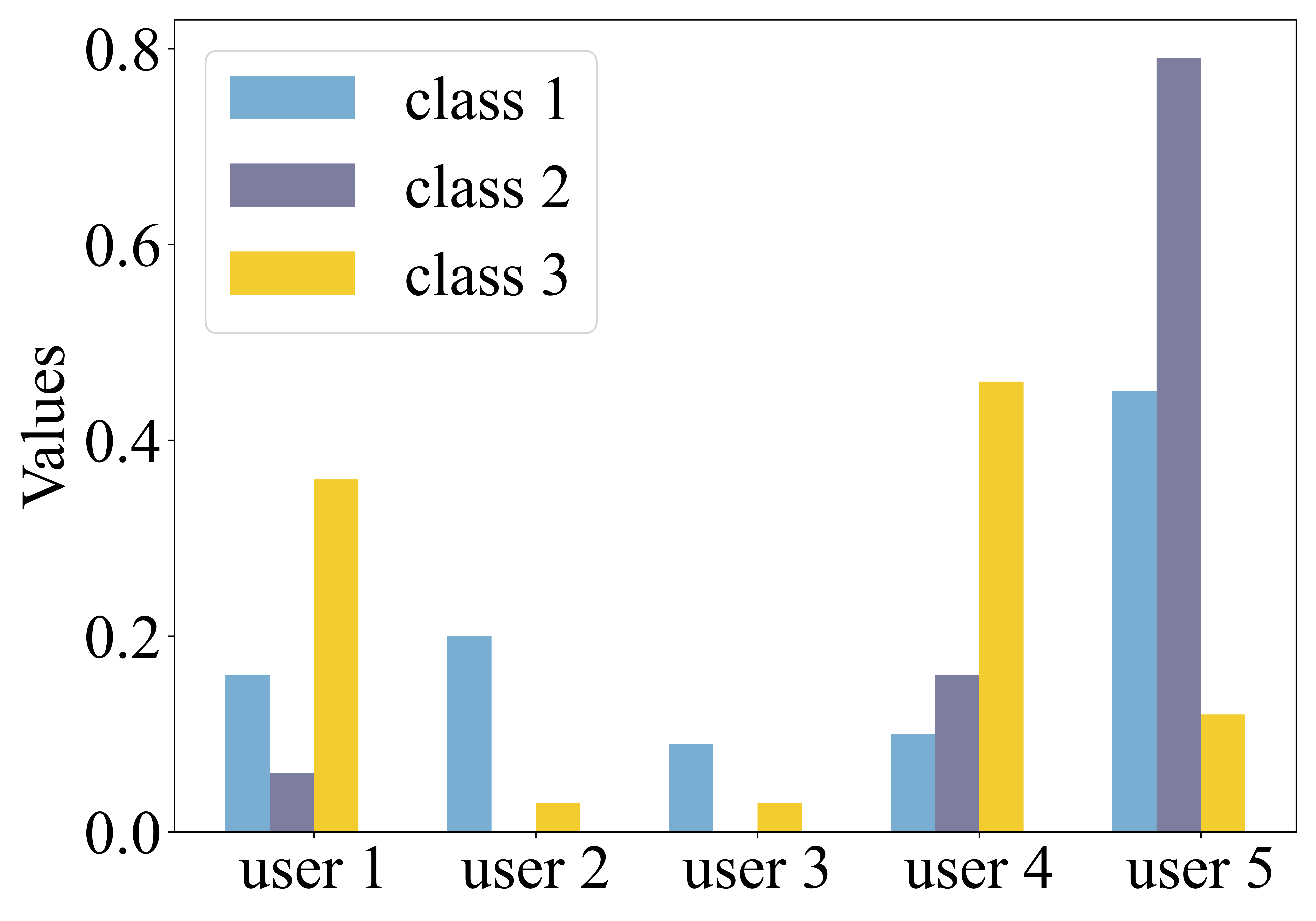}
        \caption*{\white{b}(d) CP}
    \end{minipage}
    
    \caption{Data distribution of four datasets for different clients.}
    \label{distribution}
\end{figure}

    \textbf{CP:}
    The CP data were collected from pumps at our partner factory with a sampling frequency of 5.12 kHz, including one reflux pump, one feeding pump, and one circulating pump, as shown in Fig. \ref{pumps}. There are three possible health states: one healthy state and two fault states. Vibration accelerometers were installed on both the motor side and the pump side surfaces of the machines. In subsequent experiments, the data collected by each vibration accelerometer are considered as one channel. The number of samples is 1250, with the length of samples being 2048. Data distribution of four datasets for different clients is shown in Fig. \ref{distribution}.

\begin{figure}[]
    \centering
    \includegraphics[width=.49\textwidth]{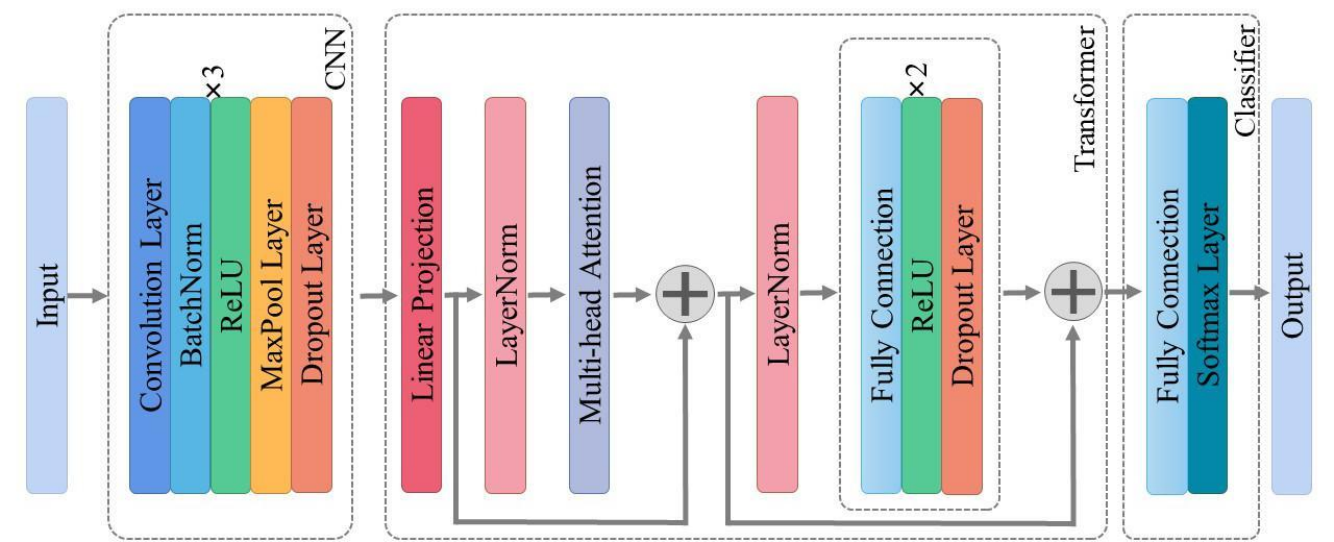}
    \vspace{-8pt}
    \caption{\textrm{Diagram of the local user's model: CNN and Transformer are served as feature extraction backbones, followed by a classifier.
    }}
        \label{basemodel}
  \end{figure}

\subsubsection{Baselines}

    To evaluate the SSFL-DCSL framework's efficacy, we compare it with several methods, including one supervised and six semi-supervised techniques, featuring two state-of-the-art semi-supervised methods:
    1) \textbf{FedAvg-Supervised}: a supervised approach based on the federal average algorithm\cite{fedavg}. 
    2) \textbf{FedAvg-FixMatch}: naive combinations of FedAvg with FixMatch\cite{FixMatch}. 
    3) \textbf{FedProx-FixMatch}: naive combinations of FedProx\cite{fedprox} with FixMatch. 
    4) \textbf{FedAvg-UDA}: naive combinations of FedAvg with unsupervised data augmentation (UDA)\cite{UDA}. 
    5) \textbf{FedProx-UDA}: naive combinations of FedAvg with UDA. 
    6) \textbf{Fed-SSMPN}: A method based on the centralized semi-supervised IFD approach SSMPN\cite{ZhangSHHW22} is extended to a federated architecture.
    7) \textbf{FedCon}\cite{fedcon}: FedAvg-FixMatch with group normalization and group averaging.

    \begin{table}[]
    \centering
    \caption{SYSTEM SETTINGS}
    \resizebox{\columnwidth}{!}{
    \begin{tabular}{c|c|c}
    \hline
    \textbf{Module} & \textbf{Parameter} & \textbf{Value/Configuration} \\
    \hline
    \multirow{7}{*}{CNN 1} & Input Channels & 3(CP) / 1(Other Datasets) \\
     & Output Channels & 32 \\
     & Kernel Size & 8x8 \\
     & Stride & 1 \\
     & Padding & 4 \\
     & Activation Function & ReLU \\
     & Pooling Layer & Max pooling, kernel size 2, stride 2 \\
     & Dropout & 0.35 \\
    \hline
    \multirow{7}{*}{CNN 2} & Input Channels & 32 \\
     & Output Channels & 64 \\
     & Kernel Size & 8x8 \\
     & Stride & 1 \\
     & Padding & 4 \\
     & Activation Function & ReLU \\
     & Pooling Layer & Max pooling, kernel size 2, stride 2 \\
    \hline
    \multirow{7}{*}{CNN 3} & Input Channels & 64 \\
     & Output Channels & 256 \\
     & Kernel Size & 8x8 \\
     & Stride & 1 \\
     & Padding & 4 \\
     & Activation Function & ReLU \\
     & Pooling Layer & Max pooling, kernel size 2, stride 2 \\
    \hline
    \multirow{3}{*}{Transformer} & Depth (Layers) & 4 \\
     & Number of Attention Heads & 4 \\
     & Hidden Dimension & 128 \\
    \hline
    \multirow{3}{*}{Projection Head} & Input Dimension & 128 \\
     & First Layer Output Dimension & 128 \\
     & Second Layer Output Dimension & 64 \\
    \hline
    \multirow{2}{*}{Classification Layer} & Input Dimension & 64 \\
     & Output Dimension & Number of Classes \\
    \hline
    \multirow{1}{*}{Dropout} & Dropout Rate & 0.35 \\
    \hline
    \end{tabular}
    }
    
    \label{systemsetting}
    \end{table}
    
\subsubsection{System Setting}
   The model comprises a feature extractor ${{f}}({{\boldsymbol{\vartheta}}})$ and a classifier ${g}({{\boldsymbol{\varphi}}})$, as shown in Fig. \ref{basemodel}. Specifically, ${{f}}({{\boldsymbol{\vartheta}}})$ consists of a three-layer CNN and a Transformer encoder. The detailed system settings are listed in Table \ref{systemsetting}. Specifically, the CNN module consists of three layers, each using an 8×8 kernel with a stride of 1. The number of channels increases from the input to 32, then to 64, and finally to 256, with each layer followed by batch normalization, ReLU activation, max pooling, and dropout. The extracted features are normalized and reshaped for the Transformer, which has 4 layers, 4 attention heads, and a hidden dimension of 128. Afterward, a projection head and fully connected layer produce the final predictions. Dropout is applied to enhance generalization.

\subsubsection{Implementation Details}\label{labelrate}
  We utilize the Dirichlet distribution Dir($\nu$) with $\nu = 0.5$ to generate non-IID data partitions on 5 clients across three datasets, resulting in diverse class and sample distributions. Each client's data is divided into training set and testing set with a 4:1 ratio. $N^{\rm l}$ and $N^{\rm u}$ represent the number of labeled and unlabeled training samples, respectively, with a label rate $\chi = N^{\rm l} / (N^{\rm l}+N^{\rm u})$. We set $\chi$ at 10\%, 20\%, and 40\% in our experiments to assess the accuracy of the model under different levels of label scarcity.  Consistent model architectures and settings are used for fair comparison, with a batch size of 16 and an Adam optimizer at a 0.001 learning rate. The PU dataset is trained for 300 rounds, while others are trained for 100 rounds. Each local training session includes one epoch. Results are averaged over five trials to reduce randomness.

\subsection{Experiment Results}
    \begin{table*}[]
    \centering 
    \caption{RESULTS FROM EXPERIMENTS ON THREE DATASETS}
    \renewcommand{\arraystretch}{1.5} 
    \resizebox{\textwidth}{!}{%
        \normalsize %
        \begin{tabular}{c|ccc|ccc|ccc} 
            \hline\hline
                Dataset    & \multicolumn{3}{c|}{PU}    & \multicolumn{3}{c|}{MFPT}  & \multicolumn{3}{c}{CWRU}   \\ \hline
            Label rate                             & 10\%           & 20\%           & 40\%           & 10\%             & 20\%           & 40\%           & 10\%           &20\%            &40\%  \\ \hline
            FedAvg-Supervised                      &78.58 $\pm2.36$ &81.17 $\pm3.32$ &86.58 $\pm4.43$ &68.42 $\pm6.35$   &80.14 $\pm0.55$ &87.52 $\pm1.99$ &66.06 $\pm3.92$ &83.39 $\pm2.70$ &89.79$\pm3.76$\\\hline
            FedAvg-FixMatch                        &79.03 $\pm1.60$ &88.68 $\pm4.45$ &92.28 $\pm2.83$ &71.76 $\pm5.76$   &85.84 $\pm2.26$ &89.77 $\pm3.20$ &67.59 $\pm4.79$ &88.51 $\pm5.81$ &95.41$\pm3.12$\\ \hline
            FedProx-FixMatch                       &80.02 $\pm0.65$ &87.27 $\pm6.90$ &96.92 $\pm0.46$ &73.09 $\pm5.31$   &81.99 $\pm6.34$ &87.74 $\pm2.01$ &73.68 $\pm4.58$ &86.50 $\pm3.03$ &91.18$\pm4.69$\\ \hline
            FedAvg-UDA                             &80.04 $\pm1.68$ &81.28 $\pm4.40$ &96.53 $\pm1.09$ &82.30 $\pm2.70$   &89.08 $\pm3.70$ &91.38 $\pm2.59$ &69.15 $\pm5.82$ &87.88 $\pm5.11$ &89.16$\pm6.00$ \\ \hline
                FedProx-UDA                            &79.28 $\pm0.95$ &83.17 $\pm3.11$ &93.98 $\pm5.30$ &81.84 $\pm6.14$   &83.38 $\pm3.35$ &87.69 $\pm2.07$ &69.87 $\pm8.35$ &86.13 $\pm3.15$ &93.06$\pm4.35$ \\ \hline
                Fed-SSMPN \cite{ZhangSHHW22}           &80.43 $\pm2.61$ &83.62 $\pm2.11$ &95.23 $\pm1.87$ &81.61 $\pm10.33$  &90.22 $\pm3.22$ &92.20 $\pm2.17$ &74.87 $\pm0.85$ &84.53 $\pm3.16$ &92.85$\pm5.08$ \\ \hline
                FedCon \cite{fedcon}                   &90.59 $\pm1.74$ &91.70 $\pm6.05$ &95.06 $\pm3.83$ &88.92 $\pm1.68$   &91.23 $\pm1.01$ &94.33 $\pm3.33$ &82.49 $\pm4.47$ &92.04 $\pm0.40$ &95.94$\pm1.19$ \\ \hline
                FedCD \cite{liu2024fedcd}              &88.72 $\pm1.47$ &92.33 $\pm1.49$ &94.18 $\pm3.51$ &89.99 $\pm3.57$   &92.38 $\pm1.19$ &95.36 $\pm3.20$ &89.19 $\pm3.16$ &90.49 $\pm1.93$ &94.19$\pm1.05$ \\ \hline
            Proposed SSFL-DCSL                          &\textbf{93.23} $\pm1.96$ &\textbf{93.42} $\pm1.08$ &\textbf{97.59}  $\pm0.83$ &\textbf{93.05} $\pm2.13$ &\textbf{96.35} $\pm1.44$ &\textbf{98.62} $\pm0.90$ &\textbf{91.50} $\pm2.78$ &\textbf{94.15} $\pm1.44$ &\textbf{96.71}$\pm0.79$ \\ \hline\hline
        \end{tabular}%
    }
    \label{tab2}
    \end{table*}

    \begin{table*}[]
    \centering 
    \large
    \caption{ABLATION RESULTS FROM EXPERIMENTS ON THREE DATASETS}
    \renewcommand{\arraystretch}{1.5} 
    \resizebox{\textwidth}{!}{%
        \normalsize %
        \begin{tabular}{c|ccc|ccc|ccc} 
            \hline\hline
                Dataset    & \multicolumn{3}{c|}{PU}    & \multicolumn{3}{c|}{MFPT}  & \multicolumn{3}{c}{CWRU}   \\ \hline
            Label rate                  & 10\%           & 20\%           & 40\%           & 10\%             & 20\%           & 40\%           & 10\%           &20\%            &40\%  \\ \hline
            FedAvg-Supervised           &78.58 $\pm2.36$ &81.17 $\pm3.32$ &86.58 $\pm4.43$ &68.42 $\pm6.35$   &80.14 $\pm0.55$ &87.52 $\pm1.99$ &66.06 $\pm3.92$ &83.39 $\pm2.70$ &89.79$\pm3.76$\\\hline
            PTA                         &80.39 $\pm0.95$ &90.32 $\pm1.31$ &92.15 $\pm2.17$ &72.49 $\pm4.97$   &86.61 $\pm6.93$ &89.33 $\pm3.45$ &78.56 $\pm5.61$ &85.93 $\pm1.63$ &90.88$\pm1.91$\\ \hline
            PTA+LCL(Naive)              &83.78 $\pm1.71$ &91.77 $\pm2.09$ &95.89 $\pm1.87$ &78.47 $\pm1.50$   &90.11 $\pm4.20$ &91.15 $\pm5.37$ &87.82 $\pm0.80$ &92.27 $\pm1.02$ &95.74$\pm0.61$\\ \hline
            PTA+GCL+LCL(Naive)          &88.72 $\pm2.38$ &91.95 $\pm0.97$ &96.99 $\pm0.56$ &91.50 $\pm1.01$   &91.62 $\pm0.24$ &95.92 $\pm0.69$ &88.89 $\pm0.72$ &92.57 $\pm1.44$ &96.41$\pm0.16$ \\ \hline
            PTA+GCL+TLAW+LCL(Naive)     &92.59 $\pm0.61$ &93.08 $\pm0.87$ &97.51 $\pm0.32$ &92.69 $\pm0.59$   &93.02 $\pm0.65$ &96.69 $\pm0.85$ &90.34 $\pm0.60$ &93.64 $\pm0.34$ &96.44$\pm0.27$\\ \hline
            PTA+GCL+TLAW+LCL(+SPNS)     &92.72 $\pm1.16$ &93.38 $\pm2.61$ &96.03 $\pm1.95$ &92.96 $\pm3.55$   &95.60 $\pm0.61$ &97.14 $\pm2.72$ &91.32 $\pm2.06$ &93.77 $\pm0.83$ &96.62$\pm0.47$ \\ \hline
            Proposed SSFL-DCSL  &\textbf{93.23} $\pm1.96$ &\textbf{93.42} $\pm1.08$ &\textbf{97.59}  $\pm0.83$ &\textbf{93.05} $\pm2.13$ &\textbf{96.35} $\pm1.44$ &\textbf{98.62} $\pm0.90$ &\textbf{91.50} $\pm2.78$ &\textbf{94.15} $\pm1.44$ &\textbf{96.71}$\pm0.79$ \\ \hline\hline
        \end{tabular}%
    }
    \label{ablation}
    \end{table*}   

    \begin{table}[]
    \centering 
    \caption{\small {RESULTS FROM EXPERIMENTS ON CP DATASET}}
    \renewcommand{\arraystretch}{1.5} 
    \footnotesize 
    \begin{tabular}{c|ccc} 
        \hline\hline
        Label rate                       & 10\%           & 20\%           & 40\%             \\ \hline
        FedAvg-Supervised                &71.99 $\pm2.17$ &74.13 $\pm1.20$ &82.85 $\pm1.77$ \\\hline
        FedAvg-FixMatch                  &74.08 $\pm3.02$ &77.74 $\pm1.88$ &88.44 $\pm3.31$ \\ \hline
        FedProx-FixMatch                 &72.85 $\pm4.46$ &79.71 $\pm3.55$ &87.31 $\pm2.60$ \\ \hline
        FedAvg-UDA                       &73.17 $\pm4.60$ &75.84 $\pm4.19$ &87.38 $\pm1.22$  \\ \hline
        FedProx-UDA                      &74.83 $\pm4.44$ &78.41 $\pm2.95$ &86.63 $\pm2.09$  \\ \hline
        Fed-SSMPN\cite{ZhangSHHW22}                        &76.54 $\pm3.88$ &84.95 $\pm3.01$ &89.76 $\pm4.71$ \\ \hline
        FedCon\cite{fedcon}                           &78.79 $\pm1.18$ &81.83 $\pm3.55$ &89.21 $\pm2.05$  \\ \hline
        FedCD\cite{liu2024fedcd}                            &79.43 $\pm4.38$ &88.80 $\pm3.65$ &90.73 $\pm2.75$   \\ \hline
        Proposed SSFL-DCSL               &\textbf{86.01} $\pm1.97$ &\textbf{89.44} $\pm2.22$ &\textbf{92.06}  $\pm1.73$ \\ \hline\hline
    \end{tabular}%
    \label{daqi}
    \end{table}

    Table \ref{tab2} and Table \ref{daqi} report the evaluation results of our method and the state-of-the-art methods on four datasets. Firstly, we can observe that our proposed SSFL-DCSL outperforms all baseline methods on the three benchmark datasets for different label rates. Second, it can be seen that all semi-supervised methods outperform supervised methods. For example, on the PU, MFPT, CWRU, and CP datasets, SSFL-DCSL achieves average accuracies of 94.75\%, 96\%, 94.12\%, and 89.12\%, respectively, while the average accuracies of the supervised methods are only 82.11\%, 78.63\%, 79.75\%, and 76.21\%. Furthermore, our proposed SSFL-DCSL, using only 10\% labeled data, achieves comparable or even superior performance to supervised training with 20\% labeled data. This demonstrates that SSFL-DCSL can effectively learn useful feature representations from unlabeled data.
    
    It is noteworthy that as the label rate increases, the performance of all methods significantly improves, indicating that the model performance is significantly influenced by the quantity of labels. When the label rate is as low as 20\%, compared to the situation with a 40\% label rate, SSFL-DCSL's accuracy on the three datasets decreases by 4.18\%, 2.27\%, 2.56\%, and 2.62\%, respectively. In contrast, the accuracy of supervised learning decreases by 5.41\%, 7.38\%, 6.4\%, and 8.72\%, respectively, indicating that the number of labels has a greater impact on supervised learning than on SSFL-DCSL. This is because the mining of unlabeled data improves the robustness of the model and reduces the dependence on a large number of labels. 

    These improvements mentioned above can be attributed to three main factors. First, the proposed method aggregates prototypes instead of models when data distributions differ, which allows knowledge sharing among clients while avoiding model bias.  Second, low-quality pseudo labels can mislead the model into learning incorrect information.  We address this by using a truncated Laplace distribution to weight the samples, allowing the model to focus more on high-quality pseudo labels to enhance its performance. Third, a large number of features in the unlabeled data are further explored through LCL and GCL, which enhance the model's representation learning capability.

\subsection{Ablation Study}
    A series of ablation experiments were conducted to evaluate the contribution of each key component in the proposed SSFL-DCSL framework, including prototype aggregation (PTA), local contrastive loss (LCL), global contrastive loss (GCL), the truncated laplace-based adaptive weighting (TLAW), as well as the enhancements in LCL, i.e., selective positive/negative sampling (SPNS) and dynamic temperature (DT). (1) PTA: Replacing the vanilla FedAvg-supervised aggregation with PTA, which aggregates prototypes rather than full models, to validate its effectiveness. (2) PTA+LCL(Naive): Adding a naive LCL (without SPNS and DT) on top of PTA to explore the benefit of contrastive learning on local features. (3) PTA+GCL+LCL(Naive): Incorporating the GCL module to further promote alignment between local and global prototypes. (4) PTA+GCL+TLAW+LCL(Naive): Extending the previous combination with TLAW to evaluate the impact of confidence-based pseudo-label weighting. (5) PTA+GCL+TLAW+LCL(+SPNS): Further adding SPNS to the LCL for more robust selection of positive and negative pairs. (6) Proposed SSFL-DCSL: Adding Dynamic Temperature (DT) to validate its role in stabilizing the contrastive training process. The results for the ablation studies across the three benchmark datasets (PU, MFPT, CWRU) and different label rates are reported in Table \ref{ablation}.

    First, it can be seen that the diagnostic accuracy of  `FedAvg-supervised' is relatively low, while  `PTA' has an average improvement of 5.51\%, 4.33\%, and 5.38\% on the PU, MFPT, and CWRU datasets, respectively. This is because  `FedAvg-supervised' directly aggregates users' models, which vary greatly due to data heterogeneity, resulting in model dispersion when directly aggregated. In  `PTA', the server aggregates prototypes rather than models, allowing clients to share more comprehensive feature representation information with other clients during personalized updates.

    Second, the introduction of LCL  showed enhancements of 3.4\%, 5.53\%, and 9.24\% on the PU, MFPT, and CWRU datasets, respectively, with a label rate of just 10\%. This highlights that the integration of LCL effectively harnesses unlabeled data for learning representations, allowing it to capture diverse and distinguishing sample characteristics, consequently boosting the model's effectiveness.

    Third, the `GCL' model has shown a 13. 03\% improvement on the MFPT dataset and a 1.07\% improvement on the CWRU dataset, both at a 10\% labeling rate. The increase in performance is notably more pronounced on the MFPT dataset, characterized by more intricate working conditions and higher learning challenges. This highlights the advantages of sharing prototype knowledge from different users in training the model for individual users. Such sharing enables the model to learn from diverse data distributions, ultimately improving its ability to generalize.

    Fourth, the integration of `TLAW' has significantly increased the effectiveness of the model. For example, when applied to the PU dataset, it increased the model's accuracy by 0.51\% to 3.87\% for various labeling rates. In particular, the enhancement is particularly noticeable when the labeling rate is set to 10\%. This is primarily due to the fact that in such challenging training scenarios, the model's performance is typically poor, leading to less reliable pseudolabels. `TLAW' plays a vital role in regulating the weights assigned to low-quality samples, thereby enhancing the overall training performance.

    \begin{figure}[]
    \centering
    \begin{minipage}{0.48\linewidth} 
        \centering
        \includegraphics[scale=0.04]{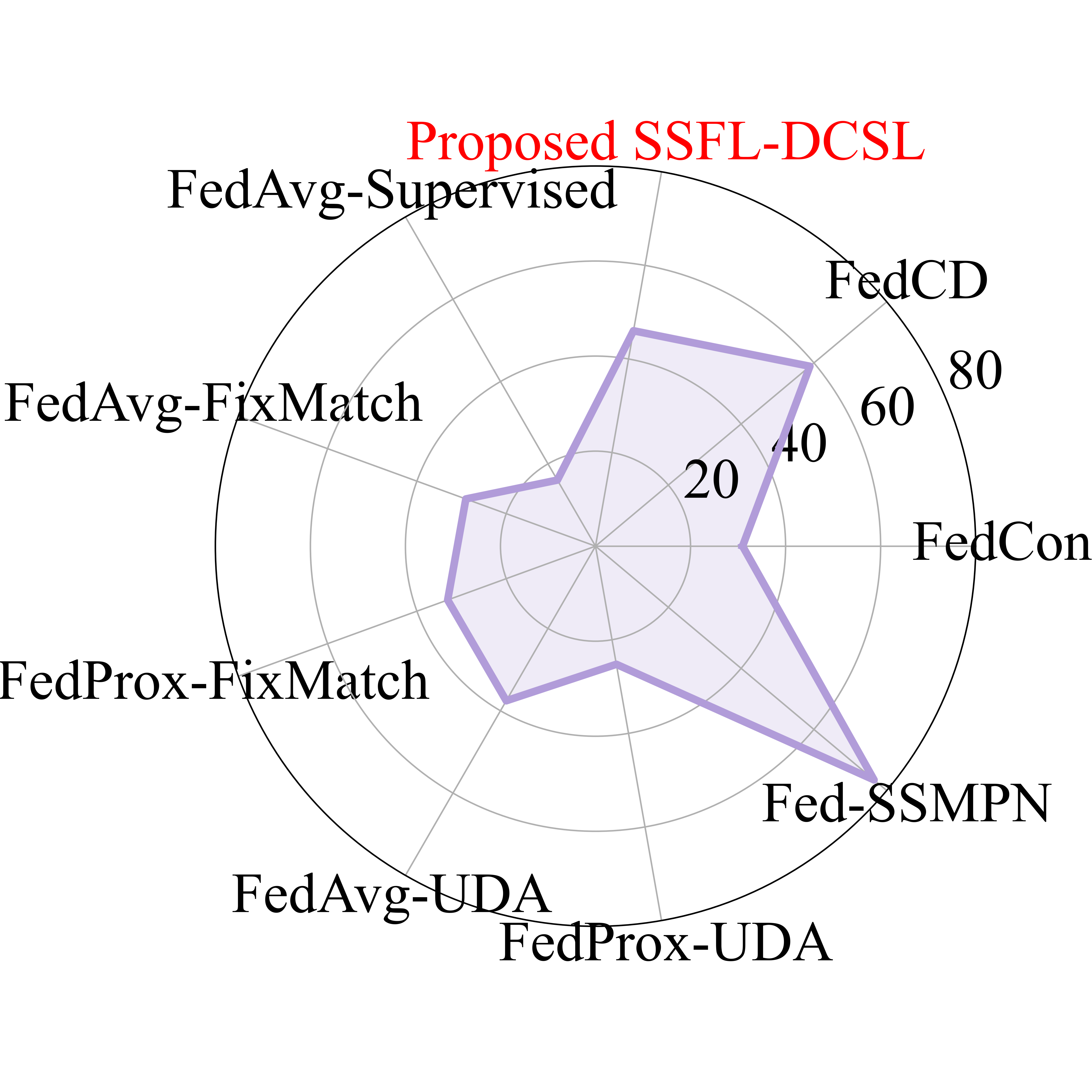}
        \vspace{-2mm}  
        \caption*{\white{b}(a) Training Time (/ms).}
    \end{minipage}%
    \hfill 
    \begin{minipage}{0.48\linewidth}
        \centering
        \includegraphics[scale=0.04]{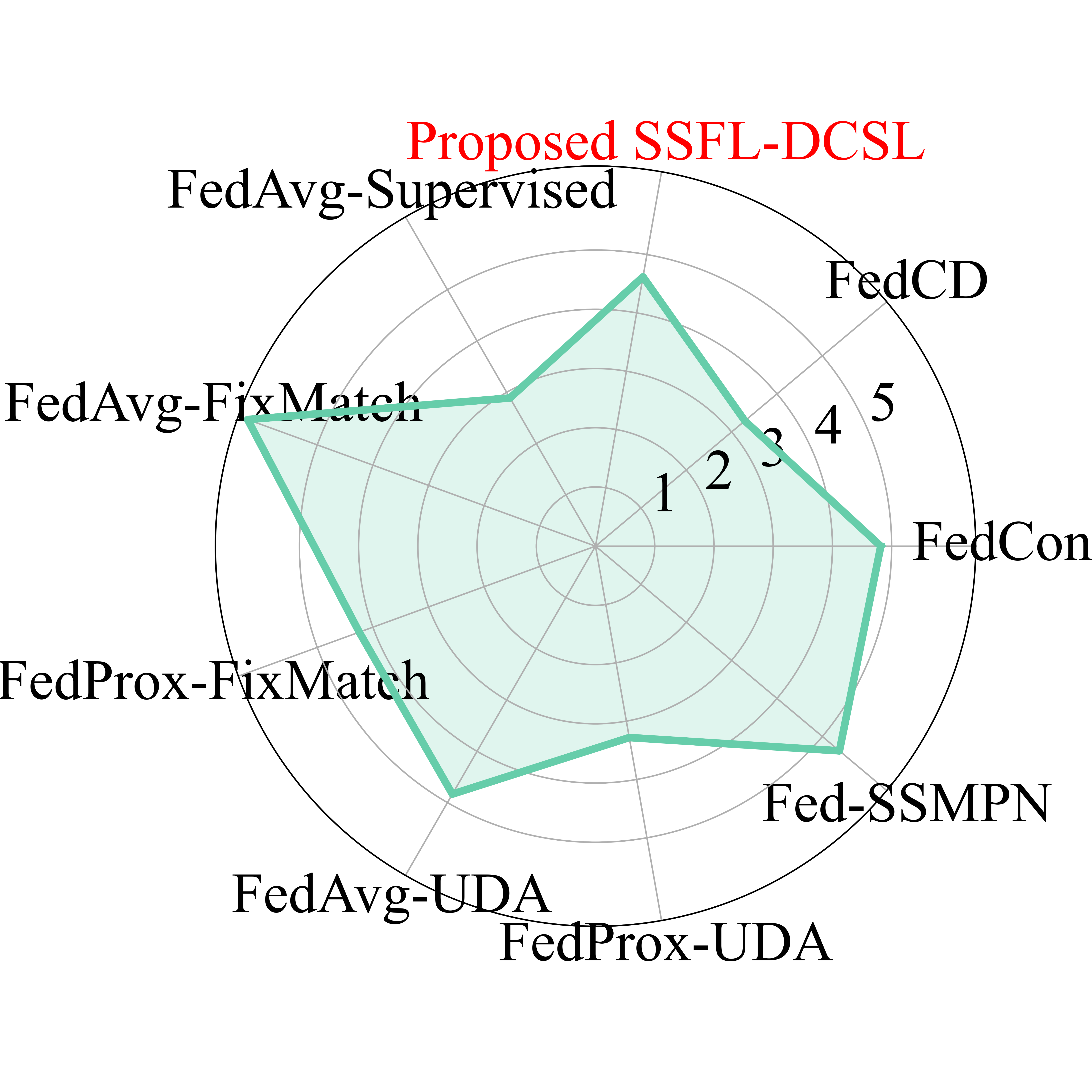}
        \vspace{-2mm}  
        \caption*{\white{b}(b) Testing Test (/ms).}
    \end{minipage}
    
    \caption{Radar chart of proposed SSFL-DCSL computational efficiency of PU dataset when $\chi=0.4$. (a) The training time of a single sample (/ms). (b) The testing time of a single sample (/ms).}
    \label{time}
    \end{figure}

    Finally, incorporating SPNS and DT into LCL yields the best results. The full model, i.e., `Proposed SSFL-DCSL', achieves the highest accuracy across all scenarios. At 10\% label rate, the proposed method reaches 93.23\%, 93.05\%, and 91.50\% on PU, MFPT, and CWRU, respectively, representing consistent and significant improvements over all ablation baselines. The successive gains from adding SPNS and DT confirm their complementary benefits. SPNS ensures more reliable pair sampling in contrastive learning, while DT adapts the contrastive temperature according to batch confidence, thereby stabilizing the optimization and further enhancing feature discriminability.

    In summary, each proposed component contributes incrementally to the overall performance, and their combination enables SSFL-DCSL to maintain high diagnostic accuracy even in scenarios with extremely limited labeled data and severe data heterogeneity. The ablation study conclusively demonstrates the necessity and effectiveness of the full design.

 \subsection{Computational Efficiency}
    Fig. \ref{time} shows radar charts illustrating the training and testing times per sample for different methods on the PU dataset at a labeling rate of 0.4. It can be observed that, first, although the fully supervised method exhibits the fastest inference speed, its reliance solely on labeled data results in the lowest accuracy. With the introduction of consistency regularization (FedAvg-FixMatch / FedProx-FixMatch), training overhead roughly doubles, yet accuracy improves substantially. Second, Fed-SSMPN and FedCD incur markedly higher training costs due to their additional contrastive learning and knowledge distillation components. Third, the proposed SSFL-DCSL method achieves the highest accuracy (97.59\%) with moderate training (46 ms/sample) and testing (5 ms/sample) latency. Compared with other methods, SSFL-DCSL strikes the best balance between accuracy and efficiency, making it well suited for deployment in real-world industrial settings.

    Moreover, SSFL-DCSL dramatically reduces communication overhead: whereas all other methods exchange the full model parameters (approximately 8.6 MB per round), the proposed approach only transmits the compact prototype representations (approximately 0.06 MB), yielding over a 99\% reduction in data transmission. This lightweight communication, combined with its superior accuracy-efficiency trade-off, makes SSFL-DCSL particularly attractive for bandwidth-constrained federated deployments in industrial environments.

\subsection{Impact of Client Dropout}

     Table \ref{drop} shows the impact of client dropout on model accuracy for the CWRU dataset with a label rate of 0.4. The results show, first, that SSFL-DCSL is highly tolerant to packet loss or temporary client drop-outs: losing just one client’s prototype lowers accuracy only slightly, from 96.71\% with full communication to 96.09\%. However, when too many prototypes are missing, global feature alignment degrades noticeably, which in turn harms classification. For example, omitting three clients’ prototypes reduces accuracy to 93.57\%. 

     The reasons for the above results are as follows: First, during prototype aggregation, the server employs a weighted averaging strategy based on the number of prototypes and uses momentum-based updates as described in (\ref{ema}). This ensures that global prototypes maintain relatively stable class centers even when some information is missing. Second, during local training, pseudo-labels with lower confidence are assigned smaller weights, reducing noise accumulation and thus allowing clients to produce relatively reliable local prototypes even under incomplete communication conditions.

     \begin{table}[!t]
    \renewcommand{\arraystretch}{1.2}
    \footnotesize
    \caption{THE ACCURACY OF THE CWRU DATASET WHEN $\chi=0.4$ FOR DIFFERENT NUMBERS OF STRAGGLER USERS}
    \centering
    \resizebox{\linewidth}{!}{
    \begin{tabular}{c|c|c|c}
    \hline\hline
    Stragglers   & 1         & 2      & 3    \\ \hline
    Acc & 96.09 $\pm$ 0.60 & 95.67 $\pm$ 0.83 & 93.57 $\pm$ 1.19 \\ \hline\hline
    \end{tabular}
    }
    \label{drop}
    \end{table}

\section{Conclusion}\label{5}
   This paper has proposed a semi-supervised FL framework, SSFL-DCSL, which integrates dual contrastive loss and soft labeling to address data and label scarcity for distributed clients with few labeled samples while safeguarding user privacy.  Specifically, first, a sample weighting function based on the Laplace distribution has been designed to alleviate bias caused by low confidence in pseudo labels during the semi-supervised training process. Second, a dual contrastive loss has been introduced to mitigate model divergence caused by different data distributions, comprising local contrastive loss and global contrastive loss. Third, local prototypes have been aggregated on the server with weighted averaging and updated with momentum to share knowledge among clients.  Ablation studies have validated the effectiveness of our methods. Despite these advancements, the lack of explainability of the model remains a limitation. Future work will focus on integrating multisensor and multimodal data to further explore the correlations between multiple variables and enhance performance.

\appendices

\section{Proof of Theorem \ref{thm:main}}\label{proof}
In this section, we will provide a detailed definition and derivation of the quantity and quality of pseudo-labels.

First, define the weight function of the sample as $\lambda ({\rm{\boldsymbol{p}}}) \in \left[ {0,{\lambda _{\max }}} \right]$ which is related to the distribution of $\boldsymbol{p}$ defined as   ${\rm{\boldsymbol{p}}} \in \left\{ {{\rm{\boldsymbol{p}}}(y|\boldsymbol{u}_i);\boldsymbol{u}_i \in {\mathcal{U}}} \right\}$. Second, the quantity $f(\boldsymbol{p})$ is defined as the ratio of unlabeled data in the weighted unsupervised loss. That is, quantity is the average sample weight of unlabeled data, given by
\begin{equation}
    f(\boldsymbol{p}) = \sum\nolimits_i^U {\frac{{\lambda ({\boldsymbol{p}_i})}}{U}}  = {\mathbb{E}_U}[\lambda ({\boldsymbol{p}_i})],
\end{equation}
where each unlabeled data is sampled evenly in ${\mathcal{U}}$.
	
The quality $g(\boldsymbol{p})$ of the pseudolabels is defined as the percentage of correct pasudo labels added to the weighted unsupervised loss, assuming that the ground truth label ${y^{\rm u}}$ for the unlabeled data is known. The 0/1 correct indicator function $\gamma (\boldsymbol{p})$ has been defined as
\begin{equation}
    \gamma (\boldsymbol{p}) = \mathds{1}(\arg \max (\boldsymbol{p}) = {y^{\rm u}}) \in \{ 0,1\}.
\end{equation}
We can formulate quality as,
\begin{equation}
    \begin{array}{l}
        g(\boldsymbol{p}) 
        = \sum\nolimits_i^U {\gamma ({\boldsymbol{p}_i})\frac{{\lambda ({\boldsymbol{p}_i})}}{{\sum\nolimits_j^U {\lambda ({\boldsymbol{p}_j})} }}} \\
        = \sum\nolimits_i^U {\gamma ({\boldsymbol{p}_i})} \bar \lambda ({\boldsymbol{p}_i})\\
        = {E_{\bar \lambda (\boldsymbol{p})}}[\gamma (\boldsymbol{p})]\\
        = {E_{\bar \lambda (\boldsymbol{p})}}[\mathds{1}(\arg \max (\boldsymbol{p}) = {y^{\rm u}})] \in [0,1],
    \end{array}
\end{equation}
where $\bar \lambda (\boldsymbol{p}) = \lambda (\boldsymbol{p})/\sum {\lambda (\boldsymbol{p})} $ is the probability mass function of $\boldsymbol{p}$ being close to ${y^{\rm u}}$.
	
In this paper, we propose a weight function based on the Laplacian mechanism to overcome the balance between pseudo-label quantity and quality. We directly model the probability mass function of $ \max(\boldsymbol{p})$ and derive the sample weight function $\lambda (\boldsymbol{p})$ from it.

We assume that the confidence of the model predictions $\max(\boldsymbol{p})$ generally follows a Laplace distribution $\mathcal{P}(\max(\boldsymbol{p});{{\hat \mu }_t},{{\hat b}_t})$ for $\max (\boldsymbol{p}) < \mu_t$ and a uniform distribution for $\max (\boldsymbol{p}) > \mu_t$, where ${b_t} = \sqrt {\frac{1}{2}\sigma _t^2} $. As the model trains, ${\mu _t}$ and ${\sigma _t}$ change. The sample weighting function is defined as
\begin{equation}
    \lambda(\boldsymbol{p}) = \begin{cases}
        \lambda_{\max} \cdot 2b_t \cdot \phi(\max({\boldsymbol{p}}; \mu_t, b_t)), & \text{if } \max(\boldsymbol{p}) < \mu_t, \\
        \lambda_{\max}, & \text{otherwise}.
    \end{cases}
\end{equation}

Without loss of generality, let us assume $\max ({\boldsymbol{p}_i}) < {\mu _t}$ for $i \in [0,\frac{U}{2}]$, as ${\mu _t} = \frac{1}{U}\sum\nolimits_i^U {\max ({\boldsymbol{p}_i})} $ and thus $\mathcal{P}(\max (\boldsymbol{p}) < {\mu _t}) = 0.5$.

Therefore, the sum of ${\lambda (\boldsymbol{p})} $ is calculated as 
\begin{equation}
    \begin{array}{l}
        \sum\nolimits_i^U {\lambda ({\boldsymbol{p}_i})} \\
        = \sum\nolimits_{i = 1}^{{\textstyle{U \over 2}}} {\lambda ({\boldsymbol{p}_i})}  + \sum\nolimits_{j = {\textstyle{U \over 2}} + 1}^U {\lambda ({\boldsymbol{p}_j})} \\
        = \sum\nolimits_{i = 1}^{{\textstyle{U \over 2}}} {{\lambda _{\max }}2{b_t}\phi (\max ({\boldsymbol{p}_i};{\mu _t},{b_t}))}  + \sum\nolimits_{j = {\textstyle{U \over 2}} + 1}^U {{\lambda _{\max }}} \\
        = {\lambda _{\max }}(\frac{U}{2} + \sum\nolimits_{i = 1}^{{\textstyle{U \over 2}}} {\exp ( - \frac{{|\max ({\boldsymbol{p}_i}) - {\mu _t}|}}{{{b_t}}})} )
    \end{array}
\end{equation}

Further, the quantity of pseudo labels is calculated as
\begin{equation}
\begin{array}{l}
    f(\boldsymbol{p})
    =\frac{1}{U}\sum\nolimits_i^U {\lambda ({\boldsymbol{p}_i})} \\
    =\frac{1}{U}(\sum\nolimits_{i = 1}^{{\textstyle{U \over 2}}} {\lambda ({\boldsymbol{p}_i})}  + \sum\nolimits_{j = {\textstyle{U \over 2}} + 1}^U {\lambda ({\boldsymbol{p}_j})} )\\
    = \frac{{{\lambda _{\max }}}}{U}(\frac{U}{2} + \sum\nolimits_{i = 1}^{{\textstyle{U \over 2}}} {\exp ( - \frac{{|\max ({\boldsymbol{p}_i}) - {\mu _t}|}}{{{b_t}}})} )\\
    = \frac{{{\lambda _{\max }}}}{2} + \frac{{{\lambda _{\max }}}}{U}\sum\nolimits_{i = 1}^{{\textstyle{U \over 2}}} {\exp ( - \frac{{|\max ({\boldsymbol{p}_i}) - {\mu _t}|}}{{{b_t}}})} .
\end{array}
\end{equation}

Since when $i \in [0,\frac{U}{2}]$, $\max (\boldsymbol{p}_i) < \mu_t$,
\begin{equation}
\begin{array}{l}
    \exp ( - \frac{{|{\textstyle{1 \over C}} - {\mu _t}|}}{{{b_t}}}) <  = \exp ( - \frac{{|\max ({\boldsymbol{p}_i}) - {\mu _t}|}}{{{b_t}}}) < 1,\\
    \frac{U}{2}\exp ( - \frac{{|{\textstyle{1 \over C}} - {\mu _t}|}}{{{b_t}}}) <  = \sum\nolimits_{i = 1}^{{\textstyle{U \over 2}}} {\exp ( - \frac{{|\max ({\boldsymbol{p}_i}) - {\mu _t}|}}{{{b_t}}})}  < \frac{U}{2},\\
    \frac{U}{2}{\lambda _{\max }} + \frac{U}{2}{\lambda _{\max }}\exp ( - \frac{{|{\textstyle{1 \over C}} - {\mu _t}|}}{{{b_t}}})\\
    <  = \frac{U}{2}{\lambda _{\max }} + {\lambda _{\max }}\sum\nolimits_{i = 1}^{{\textstyle{U \over 2}}} {\exp ( - \frac{{|\max ({\boldsymbol{p}_i}) - {\mu _t}|}}{{{b_t}}})}  < U{\lambda _{\max }},\\
    \frac{{{\lambda _{\max }}}}{2}(1 + \exp ( - \frac{{|{\textstyle{1 \over C}} - {\mu _t}|}}{{{b_t}}}))\\
    <  = \frac{{{\lambda _{\max }}}}{2} + \frac{{{\lambda _{\max }}}}{U}\sum\nolimits_{i = 1}^{{\textstyle{U \over 2}}} {\exp ( - \frac{{|\max ({\boldsymbol{p}_i}) - {\mu _t}|}}{{{b_t}}})}  < {\lambda _{\max }},\\
    \frac{{{\lambda _{\max }}}}{2}(1 + \exp ( - \frac{{|{\textstyle{1 \over C}} - {\mu _t}|}}{{{b_t}}})) <  = f(\boldsymbol{p}) < {\lambda _{\max }}.
\end{array}
\end{equation}

Hence, $f(\boldsymbol{p})$ is bounded as $\frac{{{\lambda _{\max }}}}{2} < f(\boldsymbol{p}) < {\lambda _{\max }}$.

Next, we can derive the probability mass function $\bar \lambda (\boldsymbol{p})$ on the basis of $\sum {\lambda (\boldsymbol{p})} $ and $\bar \lambda (\boldsymbol{p}) = \lambda (\boldsymbol{p})/\sum {\lambda (\boldsymbol{p})} $:
    \begin{equation}
    \bar \lambda (\boldsymbol{p}) = \begin{cases}
        \frac{{\exp ( - \frac{{|\max ({\boldsymbol{p}_i}) - {\mu _t}|}}{{{b_t}}})}}{{{\textstyle{U \over 2}} + \sum\nolimits_{i = 1}^{{\textstyle{U \over 2}}} {\exp ( - \frac{{|\max ({\boldsymbol{p}_i}) - {\mu _t}|}}{{{b_t}}})} }}, & \max ({\boldsymbol{p}_i}) < {\mu _t}, \\
        \frac{1}{{{\textstyle{U \over 2}} + \sum\nolimits_{i = 1}^{{\textstyle{U \over 2}}} {\exp ( - \frac{{|\max ({\boldsymbol{p}_i}) - {\mu _t}|}}{{{b_t}}})} }}, & \max ({\boldsymbol{p}_i}) > {\mu _t}.
    \end{cases}
\end{equation}

The quality of the pseudo labels is given by
\begin{equation}
    \begin{array}{l}
        g(\boldsymbol{p}) = \mathds{1}({{\hat {\boldsymbol{p}}}_i} = {{\boldsymbol y}^{\rm u}})\bar \lambda ({\boldsymbol{p}})\\
        = \frac{1}{{\sum\nolimits_k^U {\lambda ({{\boldsymbol{p}}_k})} }}\sum\nolimits_i^U {\gamma ({{\boldsymbol{p}}_i})} \lambda ({\boldsymbol{p}_i})\\
        = \frac{1}{{\sum\nolimits_k^U {\lambda ({{\boldsymbol{p}}_k})} }}(\sum\nolimits_{i = 1}^{{\textstyle{U \over 2}}} {\gamma ({{\boldsymbol{p}}_i})} \lambda ({{\boldsymbol{p}}_i}) + \sum\nolimits_{j = {\textstyle{U \over 2}} + 1}^U {\gamma ({{\boldsymbol{p}}_i})} \lambda ({{\boldsymbol{p}}_i}))\\
        = \sum\nolimits_i^{{\textstyle{U \over 2}}} {\gamma ({{\boldsymbol{p}}_i})\frac{{{\lambda _{\max }}2{b_t}\phi (\max ({{\boldsymbol{p}}_i});{\mu _t},{b_t})}}{{\sum\nolimits_k^U {\lambda ({p_k})} }}}  + \sum\nolimits_j^U {\gamma ({{\boldsymbol{p}}_i})} \frac{{{\lambda _{\max }}}}{{\sum\nolimits_k^U {\lambda ({p_k})} }}\\
         \ge  \sum\nolimits_i^{U - \hat U} {\frac{{{\mathds{1}}({{\hat {\boldsymbol{p}}}_i} = {\boldsymbol y}_i^{\rm u})\exp ( - \frac{{|\max ({{\boldsymbol{p}}_i}) - {\mu _t}|}}{{{b_t}}})}}{{2(U - \hat U)}}}  + \sum\nolimits_j^{\hat U} {\frac{{\mathds{1}({{\hat {\boldsymbol{p}}}_i} = {\boldsymbol y}_j^{\rm u})}}{{2\hat U}}} 
    \end{array}
\end{equation}
where $\hat U = \sum\nolimits_i^U {\mathds{1}(\max ({\boldsymbol{p}_i})}  \ge {\mu _t})$, and $U  >  {\hat U}$. Hence, we have $\sum\nolimits_i^{U - \hat U} {\frac{{{\mathds{1}}({{\hat {\boldsymbol{p}}}_i} = {\boldsymbol y}_i^{\rm u})\exp ( - \frac{{|\max ({{\boldsymbol{p}}_i}) - {\mu _t}|}}{{{b_t}}})}}{{2(U - \hat U)}}} >0 $.

Thus, the lower bound of $g(\boldsymbol{p}) $ is $\sum\nolimits_j^{\hat U} {\frac{{\mathds{1}({{\hat {\boldsymbol{p}}}_i} = {\boldsymbol y}_j^{\rm u})}}{{2\hat U}}}$.

\bibliographystyle{IEEEtran}
\bibliography{IEEEabrv,Ref}

\begin{thebibliography}{10}
\providecommand{\url}[1]{#1}
\csname url@samestyle\endcsname
\providecommand{\newblock}{\relax}
\providecommand{\bibinfo}[2]{#2}
\providecommand{\BIBentrySTDinterwordspacing}{\spaceskip=0pt\relax}
\providecommand{\BIBentryALTinterwordstretchfactor}{4}
\providecommand{\BIBentryALTinterwordspacing}{\spaceskip=\fontdimen2\font plus
\BIBentryALTinterwordstretchfactor\fontdimen3\font minus \fontdimen4\font\relax}
\providecommand{\BIBforeignlanguage}[2]{{%
\expandafter\ifx\csname l@#1\endcsname\relax
\typeout{** WARNING: IEEEtran.bst: No hyphenation pattern has been}%
\typeout{** loaded for the language `#1'. Using the pattern for}%
\typeout{** the default language instead.}%
\else
\language=\csname l@#1\endcsname
\fi
#2}}
\providecommand{\BIBdecl}{\relax}
\BIBdecl

\bibitem{iot9745085}
Y.~Chi, Y.~Dong, Z.~J. Wang, F.~R. Yu, and V.~C.~M. Leung, ``Knowledge-based fault diagnosis in industrial internet of things: A survey,'' \emph{{IEEE} Internet Things J.}, vol.~9, no.~15, pp. 12\,886--12\,900, Mar., 2022.

\bibitem{9796011}
W.~Du, P.~Hu, H.~Wang, X.~Gong, and S.~Wang, ``Fault diagnosis of rotating machinery based on 1d–2d joint convolution neural network,'' \emph{{IEEE} Trans. Ind. Electron.}, vol.~70, no.~5, pp. 5277--5285, Jun., 2023.

\bibitem{9994749}
J.~Jiao, H.~Li, and J.~Lin, ``Self-training reinforced adversarial adaptation for machine fault diagnosis,'' \emph{{IEEE} Trans. Ind. Electron.}, vol.~70, no.~11, pp. 11\,649--11\,658, Dec., 2023.

\bibitem{10043805}
B.~Hou, D.~Wang, Z.~Peng, and K.-L. Tsui, ``Adaptive fault components extraction by using an optimized weights spectrum based index for machinery fault diagnosis,'' \emph{{IEEE} Trans. Ind. Electron.}, vol.~71, no.~1, pp. 985--995, Feb., 2024.

\bibitem{10287861}
P.~Luo, Z.~Yin, D.~Yuan, F.~Gao, and J.~Liu, ``A novel generative adversarial networks via music theory knowledge for early fault intelligent diagnosis of motor bearings,'' \emph{{IEEE} Trans. Ind. Electron.}, vol.~71, no.~8, pp. 9777--9788, Oct., 2024.

\bibitem{DBLP:journals/ieeejas/YangLLLN24}
B.~Yang, Y.~Lei, X.~Li, N.~Li, and A.~K. Nandi, ``Label recovery and trajectory designable network for transfer fault diagnosis of machines with incorrect annotation,'' \emph{{IEEE} {CAA} J. Autom. Sinica}, vol.~11, no.~4, pp. 932--945, Apr., 2024.

\bibitem{DBLP:journals/ieeejas/RenWZYCN24}
J.~Ren, J.~Wen, Z.~Zhao, R.~Yan, X.~Chen, and A.~K. Nandi, ``Uncertainty-aware deep learning: {A} promising tool for trustworthy fault diagnosis,'' \emph{{IEEE} {CAA} J. Autom. Sinica}, vol.~11, no.~6, pp. 1317--1330, May., 2024.

\bibitem{fedavg}
B.~McMahan, E.~Moore, D.~Ramage, S.~Hampson, and B.~A. y~Arcas, ``Communication-efficient learning of deep networks from decentralized data,'' in \emph{Int. Conf. Artif. Intell. Stat.}, vol.~54, Apr., 2017, pp. 1273--1282.

\bibitem{ChenLYYG23}
J.~Chen, C.~Lin, B.~Yao, L.~Yang, and H.~Ge, ``Intelligent fault diagnosis of rolling bearings with low-quality data: {A} feature significance and diversity learning method,'' \emph{Reliab. Eng. Syst. Saf.}, vol. 237, p. 109343, Sep., 2023.

\bibitem{DBLP:journals/tits/ZhangLSDNTWH24}
S.~Zhang, J.~Li, L.~Shi, M.~Ding, D.~C. Nguyen, W.~Tan, J.~Weng, and Z.~Han, ``Federated learning in intelligent transportation systems: Recent applications and open problems,'' \emph{{IEEE} Trans. Intell. Transp. Syst.}, vol.~25, no.~5, pp. 3259--3285, May., 2024.

\bibitem{iot9718548}
Y.~Li, Y.~Chen, K.~Zhu, C.~Bai, and J.~Zhang, ``An effective federated learning verification strategy and its applications for fault diagnosis in industrial iot systems,'' \emph{{IEEE} Internet Things J.}, vol.~9, no.~18, pp. 16\,835--16\,849, Feb., 2022.

\bibitem{GengHLL22}
D.~Geng, H.~He, X.~Lan, and C.~Liu, ``Bearing fault diagnosis based on improved federated learning algorithm,'' \emph{Computing}, vol. 104, no.~1, pp. 1--19, Oct., 2022.

\bibitem{LiSWDMSHP22}
J.~Li, Y.~Shao, K.~Wei, M.~Ding, C.~Ma, L.~Shi, Z.~Han, and H.~V. Poor, ``Blockchain assisted decentralized federated learning {(BLADE-FL):} performance analysis and resource allocation,'' \emph{{IEEE} Trans. Parallel Distrib. Syst.}, vol.~33, no.~10, pp. 2401--2415, Dec., 2022.

\bibitem{10309847}
J.~Du, N.~Qin, D.~Huang, X.~Jia, and Y.~Zhang, ``Lightweight fl: A low-cost federated learning framework for mechanical fault diagnosis with training optimization and model pruning,'' \emph{IEEE Trans. Instrum. Meas.}, vol.~73, pp. 1--14, 2024.

\bibitem{DBLP:journals/corr/abs-2202-09027}
C.~Ma, J.~Li, K.~Wei, B.~Liu, M.~Ding, L.~Yuan, Z.~Han, and H.~V. Poor, ``Trusted {AI} in multi-agent systems: An overview of privacy and security for distributed learning,'' \emph{CoRR}, vol. abs/2202.09027, Aug., 2022.

\bibitem{10371403}
W.~Sun, R.~Yan, R.~Jin, R.~Zhao, and Z.~Chen, ``Fedalign: Federated model alignment via data-free knowledge distillation for machine fault diagnosis,'' \emph{IEEE Trans. Instrum. Meas.}, vol.~73, pp. 1--12, 2024.

\bibitem{iot9548946}
Z.~Zhang, C.~Guan, H.~Chen, X.~Yang, W.~Gong, and A.~Yang, ``Adaptive privacy-preserving federated learning for fault diagnosis in internet of ships,'' \emph{{IEEE} Internet Things J.}, vol.~9, no.~9, pp. 6844--6854, Sep., 2022.

\bibitem{10122855}
Y.~Yu, L.~Guo, H.~Gao, Y.~He, Z.~You, and A.~Duan, ``Fedcae: A new federated learning framework for edge-cloud collaboration based machine fault diagnosis,'' \emph{{IEEE} Trans. Ind. Electron.}, vol.~71, no.~4, pp. 4108--4119, May., 2024.

\bibitem{ChenHLLGL20}
Z.~Chen, G.~He, J.~Li, Y.~Liao, K.~C. Gryllias, and W.~Li, ``Domain adversarial transfer network for cross-domain fault diagnosis of rotary machinery,'' \emph{{IEEE} Trans. Instrum. Meas.}, vol.~69, no.~11, pp. 8702--8712, May., 2020.

\bibitem{ChenLHYCL22}
J.~Chen, J.~Li, R.~Huang, K.~Yue, Z.~Chen, and W.~Li, ``Federated transfer learning for bearing fault diagnosis with discrepancy-based weighted federated averaging,'' \emph{{IEEE} Trans. Instrum. Meas.}, vol.~71, pp. 1--11, Jun., 2022.

\bibitem{iot10669849}
L.~Cai, H.~Yin, J.~Lin, and Y.~Hu, ``Federated generalized zero-sample industrial fault diagnosis across multisource domains,'' \emph{{IEEE} Internet Things J.}, vol.~11, no.~23, pp. 38\,895--38\,906, Sep., 2024.

\bibitem{DBLP:journals/kbs/WanNLLL24}
L.~Wan, J.~Ning, Y.~Li, C.~Li, and K.~Li, ``Intelligent fault diagnosis via ring-based decentralized federated transfer learning,'' \emph{Knowl. Based Syst.}, vol. 284, p. 111288, 2024.

\bibitem{ChenTL23}
J.~Chen, J.~Tang, and W.~Li, ``Industrial edge intelligence: Federated-meta learning framework for few-shot fault diagnosis,'' \emph{{IEEE} Trans. Netw. Sci. Eng.}, vol.~10, no.~6, pp. 3561--3573, Nov., 2023.

\bibitem{RizveDRS21}
M.~N. Rizve, K.~Duarte, Y.~S. Rawat, and M.~Shah, ``In defense of pseudo-labeling: An uncertainty-aware pseudo-label selection framework for semi-supervised learning,'' in \emph{Int. Conf. Learn. Represent.}, May., 2021.

\bibitem{SajjadiJT16}
M.~Sajjadi, M.~Javanmardi, and T.~Tasdizen, ``Regularization with stochastic transformations and perturbations for deep semi-supervised learning,'' in \emph{Adv. Neural Inf. Process. Syst.}, Dec., 2016, pp. 1163--1171.

\bibitem{ZhangSHHW22}
X.~Zhang, Z.~Su, X.~Hu, Y.~Han, and S.~Wang, ``Semisupervised momentum prototype network for gearbox fault diagnosis under limited labeled samples,'' \emph{IEEE Trans. Ind. Inf.}, vol.~18, no.~9, pp. 6203--6213, Sep., 2022.

\bibitem{YangTYDZ24}
G.~Yang, H.~Tao, T.~Yu, R.~Du, and Y.~Zhong, ``Online fault diagnosis of harmonic drives using semisupervised contrastive graph generative network via multimodal data,'' \emph{{IEEE} Trans. Ind. Electron.}, vol.~71, no.~3, pp. 3055--3063, Mar., 2024.

\bibitem{9964271}
D.~Shan, C.~Cheng, L.~Li, Z.~Peng, and Q.~He, ``Semisupervised fault diagnosis of gearbox using weighted graph-based label propagation and virtual adversarial training,'' \emph{IEEE Trans. Instrum. Meas.}, vol.~72, pp. 1--11, Nov., 2023.

\bibitem{9999706}
X.~Pu and C.~Li, ``Meta-self-training based on teacher–student network for industrial label-noise fault diagnosis,'' \emph{IEEE Trans. Instrum. Meas.}, vol.~72, pp. 1--11, Dec., 2023.

\bibitem{10100629}
G.~Yang, H.~Tao, T.~Yu, R.~Du, and Y.~Zhong, ``Online fault diagnosis of harmonic drives using semisupervised contrastive graph generative network via multimodal data,'' \emph{IEEE Trans. Ind. Electron.}, vol.~71, no.~3, pp. 3055--3063, Apr., 2024.

\bibitem{FixMatch}
K.~Sohn, D.~Berthelot, C.-L. Li, Z.~Zhang, N.~Carlini, E.~D. Cubuk, A.~Kurakin, H.~Zhang, and C.~Raffel, ``Fixmatch: Simplifying semi-supervised learning with consistency and confidence,'' in \emph{Adv. Neural Inf. Process. Syst.}, vol.~33, Dec., 2020, pp. 596--608.

\bibitem{fedprox}
T.~Li, A.~K. Sahu, M.~Zaheer, M.~Sanjabi, A.~Talwalkar, and V.~Smith, ``Federated optimization in heterogeneous networks,'' in \emph{Proc. Mach. Learn. Syst.}, Mar., 2020.

\bibitem{fedcon}
Z.~Long, J.~Wang, Y.~Wang, H.~Xiao, and F.~Ma, ``Fedcon: {A} contrastive framework for federated semi-supervised learning,'' \emph{CoRR}, vol. abs/2109.04533, Sep., 2021.

\bibitem{liu2024fedcd}
Y.~Liu, H.~Wu, and J.~Qin, ``Fedcd: Federated semi-supervised learning with class awareness balance via dual teachers,'' in \emph{Proc. AAAI Conf. Artif. Intell.}, vol.~38, no.~4, Mar., 2024, pp. 3837--3845.

\bibitem{RSCFed}
X.~Liang, Y.~Lin, H.~Fu, L.~Zhu, and X.~Li, ``Rscfed: Random sampling consensus federated semi-supervised learning,'' in \emph{Proc. IEEE Comput. Soc. Conf. Comput. Vis. Pattern Recognit.}, Mar., 2022, pp. 10\,144--10\,153.

\bibitem{DBLP:conf/ijcai/Zhu0W0T0S24}
G.~Zhu, X.~Liu, X.~Wu, S.~Tang, C.~Tang, J.~Niu, and H.~Su, ``Estimating before debiasing: {A} bayesian approach to detaching prior bias in federated semi-supervised learning,'' in \emph{Int. Jt. Conf. Artif. Intell.}\hskip 1em plus 0.5em minus 0.4em\relax ijcai.org, May., 2024, pp. 2625--2633.

\bibitem{DBLP:conf/nips/LeeL0L24}
S.~Lee, T.~V. Le, J.~Shin, and S.~Lee, ``(fl)\({}^{\mbox{2}}\): Overcoming few labels in federated semi-supervised learning,'' in \emph{Proc. Annu. Conf. Neural Inf. Process. Syst.}, A.~Globersons, L.~Mackey, D.~Belgrave, A.~Fan, U.~Paquet, J.~M. Tomczak, and C.~Zhang, Eds., Oct., 2024.

\bibitem{tim9789131}
J.~Chen, J.~Li, R.~Huang, K.~Yue, Z.~Chen, and W.~Li, ``Federated transfer learning for bearing fault diagnosis with discrepancy-based weighted federated averaging,'' \emph{IEEE Trans. Instrum. Meas.}, vol.~71, pp. 1--11, Jun., 2022.

\bibitem{iot10891176}
Z.~Liu, Q.~Hao, H.~Shao, R.~Xin, and K.~Zhao, ``Federated distillation with lightweight generative adversarial network for servo motor bearing fault diagnosis in heterogeneous data,'' \emph{{IEEE} Internet Things J.}, vol.~12, no.~9, pp. 11\,744--11\,753, Feb., 2025.

\bibitem{iot10949604}
Z.~Ye, J.~Wu, X.~He, and W.~Jiang, ``A gradient alignment federated domain generalization framework for rotating machinery fault diagnosis,'' \emph{{IEEE} Internet Things J.}, pp. 1--1, Apr., 2025.

\bibitem{iot10742072}
H.~Ma, J.~Wei, G.~Zhang, Q.~Wang, X.~Kong, and J.~Du, ``Heterogeneous federated learning: Client-side collaborative update interdomain generalization method for intelligent fault diagnosis,'' \emph{{IEEE} Internet Things J.}, vol.~12, no.~5, pp. 5704--5718, Mar., 2025.

\bibitem{UDA}
Q.~Xie, Z.~Dai, E.~H. Hovy, T.~Luong, and Q.~Le, ``Unsupervised data augmentation for consistency training,'' in \emph{Adv. Neural Inf. Process. Syst.}, Dec., 2020.

\bibitem{DBLP:conf/ijcai/SongYZ0XK24}
Z.~Song, X.~Yang, Y.~Zhang, X.~Fu, Z.~Xu, and I.~King, ``A systematic survey on federated semi-supervised learning,'' in \emph{Int. Jt. Conf. Artif. Intell.}\hskip 1em plus 0.5em minus 0.4em\relax ijcai.org, Aug., 2024, pp. 8244--8252.

\bibitem{tim10203049}
Y.~Zhou, H.~Wang, G.~Wang, A.~Kumar, W.~Sun, and J.~Xiang, ``Semi-supervised multiscale permutation entropy-enhanced contrastive learning for fault diagnosis of rotating machinery,'' \emph{IEEE Trans. Instrum. Meas.}, vol.~72, pp. 1--10, Aug., 2023.

\bibitem{iot10909075}
X.~Lu, L.~Song, C.~Han, Q.~Jiang, W.~Xu, and H.~Wang, ``Semi-supervised contrastive domain adaptation network for fault diagnosis of rotating machinery under cross-working conditions,'' \emph{{IEEE} Internet Things J.}, pp. 1--1, Mar., 2025.

\bibitem{iot10944708}
Y.~Bi, R.~Fu, C.~Jiang, X.~Zhang, F.~Li, L.~Zhao, and G.~Han, ``Drsc: Dual-reweighted siamese contrastive learning network for cross-domain rotating machinery fault diagnosis with multi-source domain imbalanced data,'' \emph{{IEEE} Internet Things J.}, pp. 1--1, Mar., 2025.

\bibitem{tiiPeng2023OpenSetFD}
P.~Peng, J.~Lu, T.~Xie, S.~Tao, H.~Wang, and H.~Zhang, ``Open-set fault diagnosis via supervised contrastive learning with negative out-of-distribution data augmentation,'' \emph{IEEE Trans. Ind. Informat.}, vol.~19, Mar., 2023.

\bibitem{tii10922759}
H.~Zhang, Y.~Yao, Z.~Wang, J.~Su, M.~Li, P.~Peng, and H.~Wang, ``Class incremental fault diagnosis under limited fault data via supervised contrastive knowledge distillation,'' \emph{IEEE Trans. Ind. Informat.}, pp. 1--11, Mar., 2025.

\bibitem{tii10032199}
T.~Zhang, J.~Chen, S.~Liu, and Z.~Liu, ``Domain discrepancy-guided contrastive feature learning for few-shot industrial fault diagnosis under variable working conditions,'' \emph{IEEE Trans. Ind. Informat.}, vol.~19, no.~10, pp. 10\,277--10\,287, Oct., 2023.

\bibitem{tim10989640}
Y.~Dai, J.~Li, Z.~Mei, Y.~Ni, S.~Guo, and Z.~Li, ``Self-supervised learning for multimodal fault diagnosis with shapley-value weighted transformers,'' \emph{IEEE Trans. Instrum. Meas.}, vol.~74, pp. 1--14, May., 2025.

\bibitem{10151782}
Y.~Zhang, Z.~Liu, and Q.~Huang, ``A contrastive learning-based fault diagnosis method for rotating machinery with limited and imbalanced labels,'' \emph{IEEE Sens. J.}, vol.~23, no.~14, pp. 16\,402--16\,412, Jul., 2023.

\bibitem{lee2013pseudo}
D.-H. Lee \emph{et~al.}, ``Pseudo-label: The simple and efficient semi-supervised learning method for deep neural networks,'' in \emph{Proc. Int. Conf. Mach. Learn.}, vol.~3, no.~2, 2013, p. 896.

\bibitem{ijcai2021aug}
E.~Eldele, M.~Ragab, Z.~Chen, M.~Wu, C.~K. Kwoh, X.~Li, and C.~Guan, ``Time-series representation learning via temporal and contextual contrasting,'' in \emph{Int. Jt. Conf. Artif. Intell.}, Aug., 2021, pp. 2352--2359.

\end{thebibliography}

\begin{IEEEbiography}[{\includegraphics[width=1in,height=1.25in,clip,keepaspectratio]{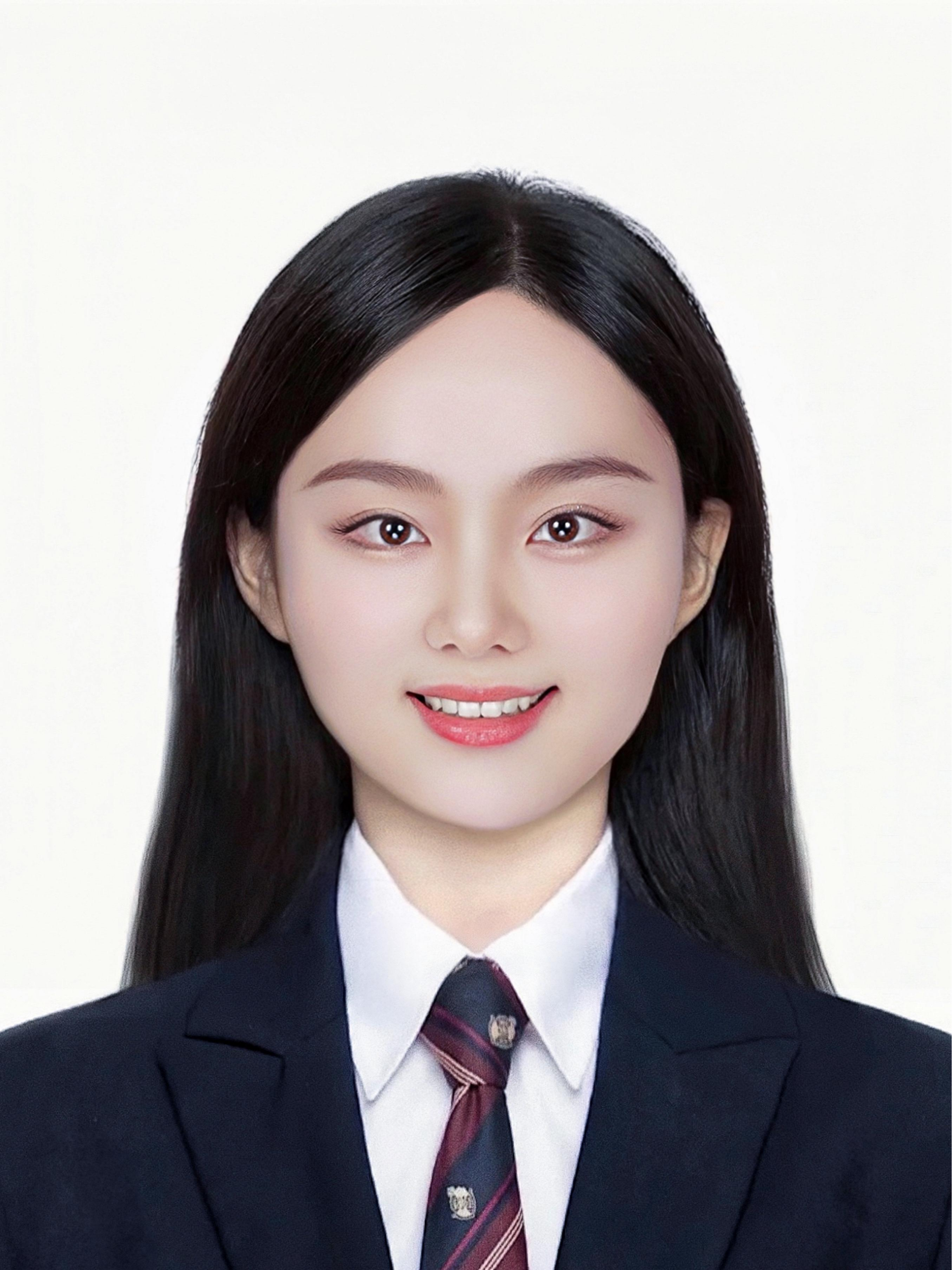}}]{Yajiao Dai} received a B.S. degree in Biomedical Engineering from Nanchang University of Nanchang, China in 2019. She is now working towards a Ph.D. degree in information and communication Engineering at the School of Electronic and Optical Engineering, Nanjing University of Science and Technology, Nanjing, Jiangsu, China. Her current research includes Federated learning and industrial AI.
\end{IEEEbiography}

\begin{IEEEbiography}[{\includegraphics[width=1in,height=1.25in,clip,keepaspectratio]{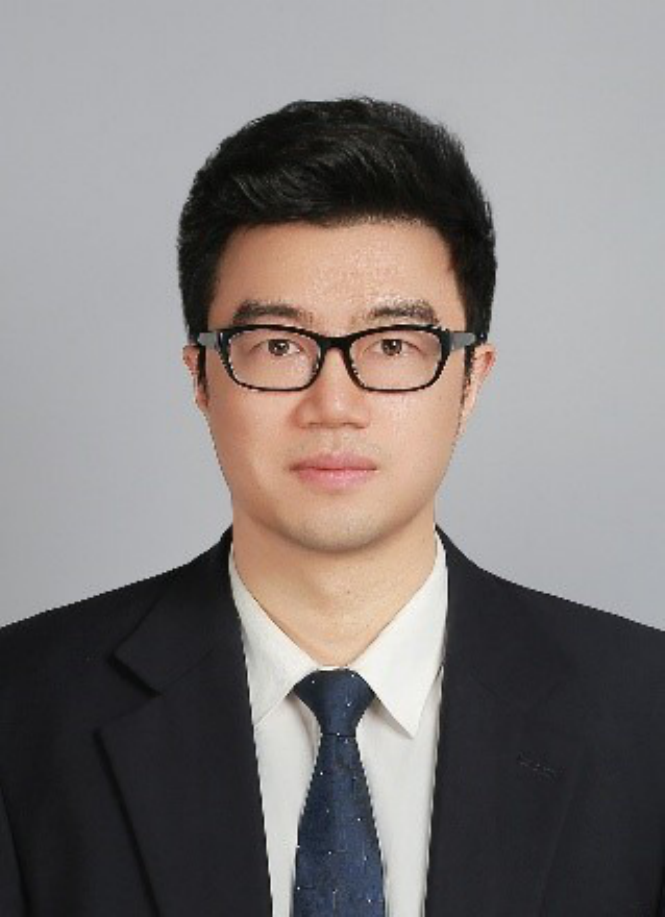}}]{Jun Li}  (M’09-SM’16-F’25) received Ph.D. degree in Electronic Engineering from Shanghai Jiao Tong University, Shanghai, P. R. China in 2009. From January 2009 to June 2009, he worked in the Department of Research and Innovation, Alcatel Lucent Shanghai Bell as a Research Scientist. From June 2009 to April 2012, he was a Postdoctoral Fellow at the School of Electrical Engineering and Telecommunications, the University of New South Wales, Australia. From April 2012 to June 2015, he was a Research Fellow at the School of Electrical Engineering, the University of Sydney, Australia. From June 2015 to June 2024, he was a Professor at the School of Electronic and Optical Engineering, Nanjing University of Science and Technology, Nanjing, China. He is now a Professor at the School of Information Science and Engineering, Southeast University, Nanjing, China. He was a visiting professor at Princeton University from 2018 to 2019. His research interests include distributed intelligence, multiple agent reinforcement learning, and their applications in ultra-dense wireless networks, mobile edge computing, network privacy and security, and industrial Internet of Things. He has co-authored more than 300 papers in IEEE journals and conferences. He was serving as an editor of IEEE Transactions on Wireless Communication and TPC member for several flagship IEEE conferences.
\end{IEEEbiography}

\begin{IEEEbiography}[{\includegraphics[width=1in,height=1.25in,clip,keepaspectratio]{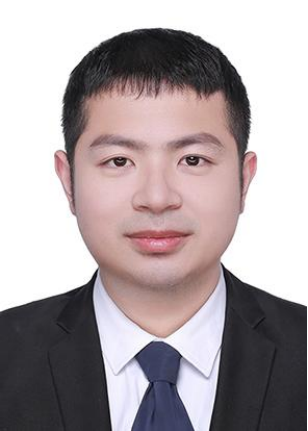}}]{Zhen Mei} (M'21) received the Ph.D. degree from Newcastle University, U.K., in 2017. He worked as a Post-Doctoral Researcher with the Singapore University of Technology and Design (SUTD) from 2017 to 2019, and a system engineer with Huawei technologies co. ltd from 2019 to 2021. He is currently an Associate Professor at Nanjing University of Science and Technology, China. His research interests include machine learning, information and coding theory for data storage and communications systems, and industrial AI. 
\end{IEEEbiography}

\begin{IEEEbiography}[{\includegraphics[width=1in,height=1.25in,clip,keepaspectratio]{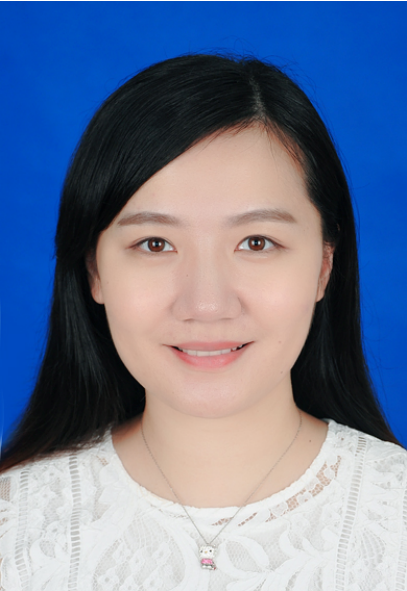}}]
 {Yiyang Ni} (Senior Member, IEEE) received the B.S. and Ph.D. degrees from the Nanjing University of Posts and Telecommunications (NJUPT), Nanjing, China, in 2008 and 2016, respectively. She is currently a Professor with Jiangsu Second Normal University, Nanjing, China. Her research interests include Massive MIMO wireless communications, intelligent surfaces and machine learning in communication systems. 
\end{IEEEbiography}

\begin{IEEEbiography}[{\includegraphics[width=1in,height=1.25in,clip,keepaspectratio]{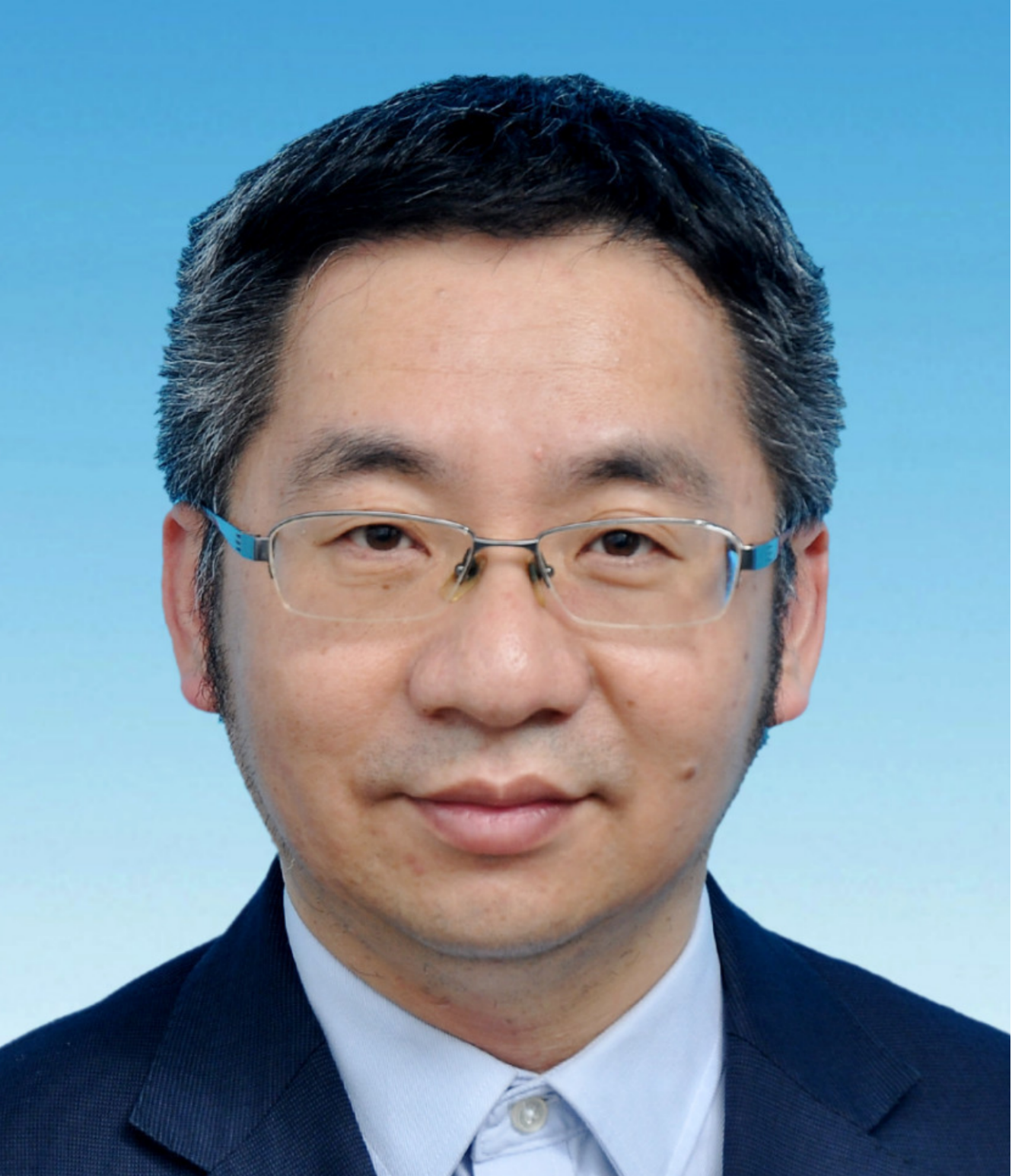}}]
 {Shi Jin} (Fellow, IEEE) received the B.S. degree in communications engineering from Guilin University of Electronic Technology, Guilin, China,in 1996, the M.S. degree from Nanjing University of Posts and Telecommunications, Nanjing, China, in 2003, and the Ph.D. degree in information and communications engineering from Southeast University, Nanjing, in 2007.From June 2007 to October 2009,he was a Research Fellow with the Adastral Park Research Campus, University College London,London, U.K. He is currently with the Faculty of the National Mobile Communications Research Laboratory, Southeast University.His research interests include wireless communications, random matrix theory, and information theory. He is serving as an Area Editor for the IEEE Transactions on Communications and IET Electronics Letters. He was an Associate Editor for the IEEE Transactions on Wireless Communications, IEEE Communications Letters, and IET Communications. Dr. Jin and his coauthors have been awarded the IEEE Communications Society Stephen O.Rice Prize Paper Award in 2011, the IEEE Jack Neubauer Memorial Award in 2023, the IEEE Marconi Prize Paper Award in Wireless Communications in 2024, and the IEEE Signal Processing Society Young Author Best Paper Award in 2010 and Best Paper Award in 2022.  
\end{IEEEbiography}

\begin{IEEEbiography}[{\includegraphics[width=1in,height=1.25in,clip,keepaspectratio]{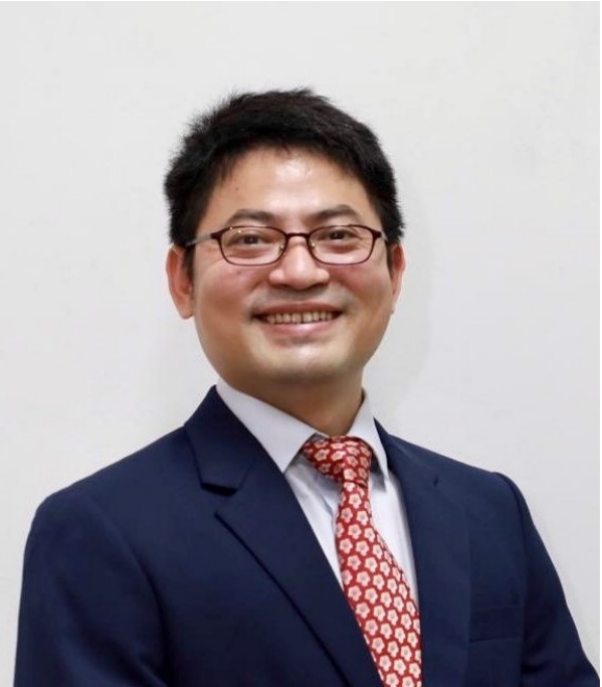}}]
 {Zengxiang Li} is Executive Vice President of
Digital Research Institute and Director of Collaborative Learning Lab of ENN Group. Dr. Li received his Ph.D. degree in the School of Computer Science and Engineering at Nanyang Technological University in 2012. His research interests include industrial IoT, blockchain, AI, federated learning, incentive mechanisms, and privacy-preserving technology.
\end{IEEEbiography}

\begin{IEEEbiography}[{\includegraphics[width=1in,height=1.25in,clip,keepaspectratio]{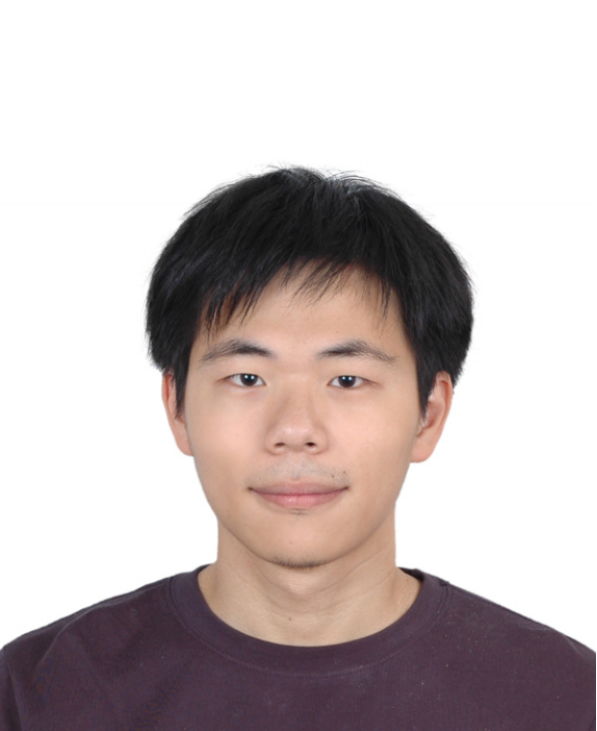}}]
 {Sheng Guo} (Member, IEEE) received his Ph.D. degree in thermal engineering from the School of Energy and Power Engineering at Huazhong University of Science and Technology, Wuhan, China, in 2020. He is currently with the Digital Research Institute of ENN Group. His research interests include IFD, deep learning, and Federated Learning.
\end{IEEEbiography}

\begin{IEEEbiography}[{\includegraphics[width=1in,height=1.25in,clip,keepaspectratio]{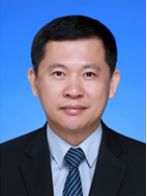}}]
 {Wei Xiang} (Senior Member, IEEE) received the B.Eng. and M.Eng. degrees in electronic engineering from the University of Electronic Science and Technology of China, Chengdu, China, in 1997 and 2000,respectively, and the Ph.D. degree in telecommunications engineering from the University of South Australia, Adelaide, Australia,in 2004. From 2004 to 2015,he was with the School of Mechanical and Electrical Engineering, University of Southern Queensland, Toowoomba, Australia. He is currently the Founding Professor and the Head of the Discipline of Internet of Things Engineering with the College of Science and Engineering, James Cook University, Cairns, Australia. He has authored or co-authored over 200 peer-reviewed journal and conference papers. His research interests are in the broad areas of communications and information theory, particularly the Internet of Things, and coding and signal processing for multimedia communications systems. He is an Elected Fellow of the IET and Engineers Australia. He received the TNO Innovation Award in 2016, and was a finalist for the 2016 Pearcey Queensland Award. He was a co-recipient of three best paper awards at 2015 WCSP, 2011 IEEE WCNC, and 2009 ICWMC. He has been awarded several prestigious fellowship titles. He was named a Queensland International Fellow (2010-2011) by the Queensland Government of Australia, an Endeavour Research Fellow(2012-2013)by the Commonwealth Government of Australia, a Smart Futures Fellow (2012-2015)by the Queensland Government of Australia, and an ISPS Invitational Fellow jointly by the Australian Academy of Science and Japanese Society for Promotion of Science (2014-2015). He is the Vice Chair of the IEEE Northern Australia Section. He was an Editor of the IEEE COMMUNICATIONS LETTERS (2015-2017), and is an Associate Editor of Telecommunications Systems (Springer). He has served in a large number of international conferences in the capacity of General Co-Chair, TPC Co-Chair, Symposium Chair, and so on.
\end{IEEEbiography}
\end{spacing}

\end{document}